\documentclass{mc}
\renewcommand\datereceived{}
\renewcommand\dateaccepted{}
\usepackage{mathrsfs}
\usepackage{caption}
\usepackage{subcaption}
\usepackage{multirow}
\usepackage{xcolor,sectsty}
\usepackage{rotating}
\definecolor{astral}{RGB}{46,116,181}
\definecolor{darkslategray}{rgb}{0.18, 0.31, 0.31}
\definecolor{warmblack}{rgb}{0.0, 0.46, 0.36}
\subsectionfont{\color{astral}}
\subsubsectionfont{\color{astral}}
\sectionfont{\color{astral}}
\usepackage{hyperref}
\usepackage{neuralnetwork}
\usepackage{ dsfont }
\usepackage{amsmath}  
\usepackage{csquotes}
\usepackage{adjustbox}
\usepackage[table]{xcolor}

\begin{document}

	\markboth{H. Pandey {\em et al.}}{An efficient wavelet-based physics-informed neural network for multiscale problems}
	\title{{\color{warmblack}An efficient wavelet-based physics-informed neural network for multiscale problems 
 }}

	\author[H. Pandey {\em et al.}]{{\bf Himanshu Pandey}\affil{1}, {\bf Anshima Singh}\affil{1,2} and {\bf Ratikanta Behera}\affil{1}\corrauth}
 
	\address{\affilnum{1}\ {Department of Computational and Data Sciences, Indian Institute of Science, Bangalore, 560012, India}\\
 \affilnum{2}\ {Department of Mathematics, University of Manchester, Oxford Road, Manchester M13 9PL, UK}}
	
	\emails{{\tt phimanshu@iisc.ac.in} (H. Pandey), {\tt anshima.singh@manchester.ac.uk} (A. Singh), {\tt ratikanta@iisc.ac.in} (R. Behera)}

%%%%% Begin Abstract %%%%%%%%%%%
\begin{abstract}
Physics-informed neural networks (PINNs) are a class of deep learning models that utilize physics in the form of differential equations to address complex problems, including those that may involve limited data availability. However, solving differential equations with rapid oscillations, steep gradients, or singular behavior becomes a challenge for PINNs. To address this, we propose an efficient {wavelet-based physics-informed neural network (W-PINN)} that learns solutions in wavelet space. Here, we represent the solution in wavelet space using a family of localized wavelets. This framework represents the solution of a differential equation with significantly fewer degrees of freedom while retaining the dynamics of complex physical phenomena. The proposed architecture enables the training process to search for solutions within the wavelet domain, where the multiscale characteristics are less pronounced compared to the physical domain. This facilitates a more efficient training for this class of problems. Furthermore, the proposed model does not rely on automatic differentiation (AD) for derivatives involved in the loss function and does not require any prior information regarding the behavior of the solution, such as the location of abrupt features. The removal of the AD requirement significantly reduces training time while maintaining accuracy. Thus, through a strategic fusion of wavelets with PINNs, W-PINNs excel at capturing localized non-linear information, making them well-suited for problems showing abrupt behavior in certain regions, such as singularly perturbed and other multiscale problems. We further analyze the convergence behavior of W-PINN through a comparative study using the Neural Tangent Kernel (NTK) theory. The efficiency and accuracy of the proposed neural network model are demonstrated in various problems, i.e., the FitzHugh-Nagumo (FHN) model, the Helmholtz equation, the Maxwell equation, the Allen-Cahn equation, and lid-driven cavity flow, along with other highly singularly perturbed non-linear differential equations. 
\end{abstract}

%%%%% Keywords %%%%%%%%%%%
	
%%%% AMS subject classifications %%%%
	
%%%% maketitle %%%%%
\maketitle
	
%%%% Start %%%%%%

\section{Introduction}\label{sec_intro}

The domain of computational science has been profoundly reshaped in recent years by the emergence of artificial intelligence. Among these developments, physics-informed neural networks (PINNs) \cite{lagaris1998artificial, MR3881695} have gained substantial recognition as a powerful alternative to traditional numerical methods for solving partial differential equations (PDEs). Traditional numerical methods such as finite difference, finite element, and finite volume approaches have long served as the workhorses of computational science. However, these techniques often struggle with complex geometries that require intricate mesh generation and face significant challenges when scaling to higher-dimensional problems. PINNs address these limitations by offering a mesh-free approach that eliminates the tedious process of mesh construction and the associated numerical artifacts that arise from poor mesh quality \cite{NN_HU2024106369}. What makes PINNs particularly compelling is their ability to generalize continuous functional representations across the entire computational domain rather than discrete solution points.

 Over the past few years, several variants of PINNs have been proposed to enhance robustness and convergence, including variational physics-informed neural networks (hp-VPINN) \cite{KHARAZMI2021113547}, gradient-enhanced physics-informed neural networks (GPINN) \cite{GPINN}, and extended physics-informed neural networks (XPINN) \cite{XPINN}, and many others to enhance the performance of conventional PINN. For more details, see a comprehensive review of PINNs  \cite{MR4457972, karniadakis2021physics} and references therein.

Despite their many significant advantages, the performance of PINNs significantly deteriorates when dealing with problems that exhibit high gradients, rapid oscillations, or singular behavior \cite{karniadakis2021physics}. Wang et al. demonstrate through extensive numerical experiments that conventional PINN methods exhibit significant limitations when applied to multiscale problems \cite{MR4267540, wang2021understanding}. Singularly perturbed differential equations represent another challenging class, as their solutions exhibit thin transition layers, often adjacent to the domain boundaries. These solutions or their derivatives undergo rapid fluctuations within specific areas while maintaining smooth behavior elsewhere \cite{MR2454024}, which presents particular difficulties for standard PINN approaches. { A key reason for these challenges is that various components (residual loss, initial condition and boundary loss) in the loss function have significantly different magnitudes and converge at different rates during training, complicating the optimization process \cite{NN_FARHANI2025106963, alkhalifa}}. Furthermore, when tackling multiscale problems characterized by multi-frequency or high-frequency features, the conventional PINN method faces dual challenges: not only does it encounter these loss-term magnitude imbalances, but it also struggles to accurately capture high-frequency functions due to spectral bias \cite{rahaman2019spectral}. These combined limitations ultimately result in compromised prediction accuracy. These limitations are also theoretically verified by Neural Tangent Kernel (NTK) theory for neural network learning, which was originally formulated by Jacot et al. \cite{NEURIPS2018_5a4be1fa}. Subsequently, Cao et al. \cite{ijcai2021p304} provided a detailed spectral analysis of NTK, and later Wang et al. \cite{MR4337814} provided a similar analysis for PINNs.
 
The recent literature has proposed several approaches to address these limitations. McClenny et. al. introduced Self-adaptive PINNs (SA-PINNs)\cite{mcclenny2023self}  that employ trainable weights to dynamically balance loss terms, while Arzani et al. \cite{MR4511357} proposed a BL-PINN approach that splits the domain into inner and outer regions to effectively handle boundary layer problems with singular perturbation. Moreover, Wang et al. developed a practical PINN framework for multiscale problems with multi-magnitude loss terms (MMPINN)\cite{wang2024practical}, incorporating specialized architectures and regularization strategies. {Complementary approaches include spectral PINNs that incorporate specialized basis functions. Notable examples are Gabor PINN\cite{GaborPINN, GaborEnhanced} and physics-informed radial basis networks (PIRBNs)\cite{PIRBN}. These methods integrate oscillatory or localized basis functions directly into the network architecture. This integration enhances the network's ability to approximate high-frequency components and localized features in the solution.} 
% {In a complementary direction, basis based spectral PINNs, such as Gabor PINN\cite{GaborPINN, GaborEnhanced} and physics-informed radial basis networks (PIRBNs)\cite{PIRBN}, enhance expressivity by embedding oscillatory or localized basis functions within the network, improving approximation of high-frequency and localized features.} 
Although these methods demonstrate improved performance, they still face computational overhead from automatic differentiation (AD) and require careful tuning of multiple hyperparameters. However, AD provides machine-precision derivatives, it comes at the cost of maintaining computation-intensive computational graphs that grow increasingly complex with network depth and order of derivatives involved.

To reduce the overhead associated with AD, several alternative non-AD approaches have been proposed. Monte Carlo approximation techniques for second-order PDE derivatives were explored by Sgouralis and Spiliopoulos \cite{MR3874585}. Navaneeth et al. developed a gradient-free learning methodology utilizing random projections, successfully modeling complex fourth-order phase field fracture scenarios \cite{MR4541847}. Other developments include using meshless radial basis functions by Xiao et al. \cite{xiao2023radial} to compute spatial derivatives, which theoretically accommodates random collocation point arrangements. One more recent development in this direction is DT-PINN \cite{sharma2022accelerated}, which achieved two to four times speedups over AD-based PINNs by eliminating AD through meshless radial basis function-finite differences (RBF-FD). These methods help mitigate computational inefficiencies and improve accuracy. However, derivative-based schemes can encounter difficulties when applied to irregular geometries, higher-order PDEs, or strongly localized features, and often require auxiliary collocation points outside the physical domain. Consequently, designing a framework that is simultaneously efficient, robust, and multiscale-aware remains an open challenge.

Building upon these foundations, the motivation for our present work is two-fold. First, we seek to overcome the challenge of multi-magnitude residuals, which limit the effectiveness of conventional PINNs in capturing sharp transitions and high-frequency features in multiscale problems. Second, we aim to reduce the significant computational overhead associated with AD techniques that dominate existing implementations. {To address these challenges simultaneously, we propose a wavelet-based physics-informed neural network (W-PINN) that reformulates physics-informed learning in a multiresolution function space, rather than modifying network architectures or loss-balancing strategies, or heuristic regularization.}

{The foundational works of Mallat \cite{mallat1989theory} and Daubechies \cite{MR1162107} on wavelet analysis, demonstrate that functions exhibiting sharp gradients, oscillations, or localized singularities, often admit compact and well-conditioned representations in wavelet space. In the context of physics-informed learning, this property is particularly advantageous, as conventional PINNs approximate solutions pointwise in the physical domain, where multiscale features lead to severe loss imbalance and slow convergence of high-frequency modes, a wavelet representation decomposes the solution into scale-separated components. As a result, localized or high-frequency structures that are difficult for neural networks to learn directly become isolated within a small subset of wavelet coefficients, enabling W-PINNs to resolve boundary layers, sharp transitions, and oscillatory features more efficiently.} Incorporation of wavelets into PINNs has been attempted in \cite{uddin2023wavelets}. However, the authors of \cite{uddin2023wavelets} focused on wavelets solely as activation functions, with the overall structure of PINNs remaining unchanged from its original introduction. Our findings show that this method also fails to satisfactorily approximate the problems under consideration.

{In this study, the proposed W-PINN employs wavelets as a solution basis, representing the unknown solution as a linear combination of multi-resolution wavelet functions whose coefficients are learned by the neural network. Unlike Gabor PINN and PIRBN, which enhance approximation while remaining physical-space and AD-dependent, this formulation shifts the learning task from pointwise solution approximation to coefficient learning in wavelet space, where multiscale features are naturally separated across resolutions.} The proposed neural architecture does not rely on the automatic diﬀerentiation to compute the derivatives involved in the loss function, thereby significantly reducing training time. In addition, this method does not require prior information about the nature of the solution, making it practical and easy to implement. We used Gaussian and Mexican hat wavelets for implementation, and their performance was compared with conventional PINN and state-of-the-art methods to validate the efficacy. Our approach demonstrates high accuracy across a range of differential equations that exhibit steep gradients, rapid oscillations, singularities, and multiscale behavior, establishing it as a robust method for these classes of problems. The key contributions of this work can be summarized as follows.
\begin{itemize}   
    \item We introduce a W-PINN framework that eliminates automatic differentiation in loss function derivative computation, significantly accelerating training while maintaining or improving solution accuracy compared to recent methods in the literature.

    \item W-PINN effectively addresses the loss balancing challenges inherent in conventional PINNs, particularly for problems that exhibit multiscale phenomena, singular behavior, or rapid oscillations in their solutions.

\item The proposed method is validated through NTK theory and various examples, including the FHN model, Helmholtz equation, Allen-Cahn equation, Maxwell equation, Lid-driven cavity flow, and various other singularly perturbed problems.
    
\end{itemize}

The organization of the remaining paper is summarized as follows: Section \ref{sec:2} initiates with the introduction of standard PINNs, a summary of NTK theory for PINNs, followed by a thorough discussion on the design and operational mechanism of W-PINNs. Section \ref{Results} presents numerical results along with a comparison with other well-known methods in the literature for various differential equations. In this section, we also justify a much faster convergence of W-PINN using NTK theory. Finally, Section \ref{sec_concl} serves as the concluding segment, offering a concise summary of the key findings, contributions, and future work along with limitations. 
%%%%%%%%%%%%%%%%%%%%%%%%%%%%%%%%%%%%%%%%%%%

\section{Methodology}\label{sec:2}

\subsection{Physics-informed neural networks (PINNs)}
A neural network is a mathematical model consisting of layers of neurons interconnected via non-linear operations. These layers include an input layer for initial data, multiple hidden layers for complex computations, and an output layer for predictions. It is widely acknowledged that with enough hidden layers and neurons per layer, a neural network can approximate any function \cite{MR1015670}. The values of neurons in each layer depend on the previous layer in the following manner:

\begin{equation}
    \begin{aligned}
        &\text{Input layer:} &\boldsymbol z^{0}&=\textbf{x}\in \mathds{R}^{n_0},\\
        &\text{Hidden layers:} &\boldsymbol z^{k}&=\sigma({\boldsymbol\omega}^{k}\boldsymbol z^{k-1}+\boldsymbol b^{k}),~1\leq k\leq L-1, \\
        &\text{Output layer:} &\boldsymbol z^L&= {\boldsymbol\omega}^{L} \boldsymbol z^{L-1}+\boldsymbol b^L\in \mathds{R}^{n_L},
    \end{aligned}
    \label{'eq:nn'}
\end{equation}

where $\boldsymbol z^{k} \in \mathds{R}^{n_k}$ is the outcome of $k$-th layer, $\boldsymbol\omega^{k}\in \mathds{R}^{n_{k}\times n_{k-1}}$, $\boldsymbol b^{k}\in \mathds{R}^{n_k}$, and $n_k$ is the number of neurons in $k^{th}$ layer. Here, $\boldsymbol\omega$ and $\boldsymbol b$ denote the weight and bias, the model parameters to be optimized. Moreover, $\sigma$ is a non-linear mapping known as an activation function. This process is widely known as feed-forward propagation. We define a loss function ($\mathcal L$) for network output, which quantifies the disparity between the network's predictions and the desired outcomes. The neural network is trained using backpropagation \cite{rumelhart1986learning}, a technique wherein optimization algorithms, such as gradient descent, are used to minimize the loss function and fine-tune the model parameters ($ \boldsymbol \omega, \boldsymbol b$). Trained parameters capture the underlying structure of the problem under consideration.   
A well-known class of neural networks is physics-informed neural networks (PINNs), which incorporate the underlying physics of the system by using a governing set of equations along with initial and boundary conditions into the loss function. It measures how much the network's predicted solution violates the differential equation at several collocation points. 

Consider a PINN model, $\hat u (\boldsymbol x, t; \boldsymbol \theta)$ that predicts the solution of a differential equation in the spatiotemporal domain, $\boldsymbol x \in \Omega$ and $t\in(0,T]$. Here  $\boldsymbol \theta$ are trainable parameters of the network, which describe the non-linear relationship $(\boldsymbol x, t) \rightarrow \hat u$ according to \ref{'eq:nn'}. A general form of a partial differential equation (PDE) can be {given by}

\begin{equation}
    \begin{cases}
        \mathscr L_{\boldsymbol x, t}[u(\boldsymbol x, t)] = f(\boldsymbol x, t), \quad \boldsymbol x \in \Omega,~t\in(0,T],\\
        \mathscr B[u(\boldsymbol x, t)] = g(\boldsymbol x,t), \quad \boldsymbol x \in \partial \Omega,~t\in(0,T],\\
        \mathscr I[u(\boldsymbol x, 0)] = h(\boldsymbol x), \quad \boldsymbol x \in \Omega,
    \end{cases}
\end{equation}
%$(\boldsymbol x, t)$ denotes spatiotemporal {coordinates},
where $\mathscr L_{\boldsymbol x, t}[\cdot]$ represents differential operator, $\mathscr B[\cdot]$ and $\mathscr I[\cdot]$ correspond to boundary and initial conditions respectively. {Let $\{(\boldsymbol{x}_i^{r}, t_i^{r})\}_{i=1}^{N_r} \subset \Omega \times (0,T]$ denote a set of collocation points used to evaluate the governing differential operator. Similarly, let $\{(\boldsymbol{x}_i^{bc}, t_i^{bc})\}_{i=1}^{N_{bc}} \subset \partial\Omega \times (0,T]$ and $\{\boldsymbol{x}_i^{ic}\}_{i=1}^{N_{ic}} \subset \Omega$ denote the spatial collocation points corresponding to the boundary and initial conditions.} The loss function for PINN can be defined as $\mathcal L=\mathcal L_\text{res}+\mathcal L_\text{ic} + \mathcal L_\text{bc}$.
Here, 
\begin{equation}
    \begin{aligned}
        \mathcal L_{\mathrm{res}} &=
        \frac{1}{N_r}\sum_{i=1}^{N_r}
        \Big(
        \mathscr L\big[\hat u(\boldsymbol{x}_i^{r}, t_i^{r}; \boldsymbol{\theta})\big]
        - f(\boldsymbol{x}_i^{r}, t_i^{r})
        \Big)^2, \\[4pt]
        \mathcal L_{\mathrm{bc}} &=
        \frac{1}{N_{bc}}\sum_{i=1}^{N_{bc}}
        \Big(
        \mathscr B\big[\hat u(\boldsymbol{x}_i^{bc}, t_i^{bc}; \boldsymbol{\theta})\big]
        - g(\boldsymbol{x}_i^{bc}, t_i^{bc})
        \Big)^2, \\[4pt]
        \mathcal L_{\mathrm{ic}} &=
        \frac{1}{N_{ic}}\sum_{i=1}^{N_{ic}}
        \Big(
        \hat u(\boldsymbol{x}_i^{ic}, 0; \boldsymbol{\theta})
        - h(\boldsymbol{x}_i^{ic})
        \Big)^2,
    \end{aligned}
\label{eq:loss}
\end{equation}
where $\mathcal L_\text{res}$ denotes the residual mean squared error, $\mathcal L_\text{bc}$ and $\mathcal L_\text{ic}$ denote mean squared errors for boundary and initial conditions, respectively. During optimization, the derivative of the loss function with respect to the model parameters and derivatives of the network's output with respect to input parameters, exhibited in the loss function, are required. These derivatives are effectively computed using \enquote{automatic differentiation} \cite{MR3800512}. Thus, the objective of PINNs is to identify the optimal parameters that minimize the specified loss function, which encapsulates all physics-related information.
%%%%%%%%%%%%%%%%%%%%%%%%%%%%%%%%%%%%%%%%%%%%%%%%%%%%%%%%%%%%%%%%%%%%%%

\subsection {Neural tangent kernel for PINNs}

%The Neural Tangent Kernel (NTK) or The neural tangent kernel (NTK)?
The Neural Tangent Kernel (NTK) theory offers an effective mathematical formulation for analyzing the dynamics of neural network functions, $f(\boldsymbol x;\boldsymbol \theta)$, where $\boldsymbol x$ is input and $\boldsymbol \theta$ are the network's parameters. NTK was first proposed by Jacot et al. \cite{NEURIPS2018_5a4be1fa}. They introduced a kernel regression framework that allows the study of neural network training in the function space rather than in the parameter space. {Since the loss function is defined as a squared $L_2$-error between the network output and a target function, the resulting optimization problem is convex with respect to the network output $f(\boldsymbol{x})$, even though it remains highly nonconvex with respect to the parameters $\boldsymbol{\theta}$}, this makes analysis more tractable in function space. Jacot et al. demonstrated that, under training with an infinitesimally small learning rate, the network function $f(\boldsymbol x;\boldsymbol \theta)$  evolves according to the kernel gradient descent with respect to the NTK. Moreover, in the infinite-width limit, the NTK converges to a deterministic limiting kernel that remains constant throughout training.

Subsequent work by Cao et al. \cite{ijcai2021p304} provided a spectral analysis of the NTK, showing that smaller NTK eigenvalues lead to slower convergence of high-frequency components in the target function. This phenomenon, known as the ``spectral bias" of neural networks, states that neural networks tend to learn low-frequency features more efficiently than high-frequency ones. Building on this foundation, Wang et al. \cite{MR4337814} {applied NTK theory to PINNs, analyzing how the NTK spectrum governs the convergence rate of the training error.} Their analysis can be summarized as follows.
Consider a PINN $\hat u(\boldsymbol x;\boldsymbol \theta)$ that is an approximate solution of the following PDE. 
\begin{equation*}
\begin{cases}
    \mathscr{L}[u(\boldsymbol{x})] = f(\boldsymbol{x}), ~~~~ \boldsymbol{x} \in \Omega, \\
    \mathscr{B}[u(\boldsymbol{x})] = g(\boldsymbol{x}), ~~~~ \boldsymbol{x} \in \partial\Omega.
\end{cases}
\end{equation*}
Here, $\mathscr L$ and $\mathscr B$ represent differential operators and boundary conditions, respectively. For given data points $\big \{ \boldsymbol {x}_r^i,f(\boldsymbol{x}_r^i)\big \}_{i=1}^{N_r}$ and $\big \{\boldsymbol {x}_b^i,g(\boldsymbol{x}_b^i \big \}_{i=1}^{N_b}$ the dynamics of $\hat u$ and $\mathscr{L}[\hat u]$ follows.

\begin{equation}
    \begin{aligned}
        \left[ \begin{array}{c} 
        \frac{d \hat u(\boldsymbol{x}_b; \boldsymbol \theta(n))}{dn} \\ 
        \frac{d \mathscr{L}[\hat u](\boldsymbol{x}_r; \boldsymbol \theta(n))}{dn} 
        \end{array} \right] 
        &= -\begin{bmatrix} 
        \boldsymbol K_{uu}(n) & \boldsymbol K_{ur}(n) \\
        \boldsymbol K_{ru}(n) & \boldsymbol K_{rr}(n) 
        \end{bmatrix}
        \begin{bmatrix} 
        \hat u(\boldsymbol{x}_b; \boldsymbol \theta(n)) - g(\boldsymbol{x}_b) \\ 
        \mathscr{L}[\hat u](\boldsymbol{x}_r; \boldsymbol \theta(n)) - f(\boldsymbol{x}_r) 
        \end{bmatrix} \\[6pt]
        &= - \boldsymbol K(n) \begin{bmatrix} 
        \hat u(\boldsymbol{x}_b; \boldsymbol \theta(n)) - g(\mathbf{x}_b) \\ 
        \mathscr{L}[\hat u](\boldsymbol{x}_r; \boldsymbol \theta(n)) - f(\boldsymbol{x}_r) 
        \end{bmatrix},
    \end{aligned}
    \label{eq:ntk0}
\end{equation}
where $n$ is training step and $K(n)$ is NTK for PINN, whose $\{i,j\}$-th element is obtained by:

\begin{align*}
    (\boldsymbol{K}_{uu})_{ij}(n) &= \left\langle 
    \frac{\partial \hat u(\boldsymbol{x}_b^i; \boldsymbol{\theta}(n))}{\partial \boldsymbol{\theta}},\ 
    \frac{\partial \hat u(\boldsymbol{x}_b^j; \boldsymbol{\theta}(n))}{\partial \boldsymbol{\theta}} 
    \right\rangle, \\
    (\boldsymbol{K}_{ur})_{ij}(n) &= \left\langle 
    \frac{\partial \hat u(\boldsymbol{x}_b^i; \boldsymbol{\theta}(n))}{\partial \boldsymbol{\theta}},\ 
    \frac{\partial \mathscr{L}[\hat u](\boldsymbol{x}_r^j; \boldsymbol{\theta}(n))}{\partial \boldsymbol{\theta}} 
    \right\rangle, \\
    (\boldsymbol{K}_{rr})_{ij}(n) &= \left\langle 
    \frac{\partial \mathscr{L}[\hat u](\boldsymbol{x}_r^i; \boldsymbol{\theta}(n))}{\partial \boldsymbol{\theta}},\ 
    \frac{\partial \mathscr{L}[\hat u](\boldsymbol{x}_r^j; \boldsymbol{\theta}(n))}{\partial \boldsymbol{\theta}} 
    \right\rangle,
\end{align*}
where $\langle \cdot , \cdot \rangle$ represents the inner product in all parameters. Similarly to NTK theory for neural networks, Wang et al. \cite{MR4337814} establish that under certain assumptions, NTK for a PINN also remains constant during training; therefore, we approximate $\boldsymbol K(n) \approx \boldsymbol K(0)$. This assumption leads to a solution of \ref{eq:ntk0} as

\begin{equation}
\begin{aligned}
    &\begin{bmatrix} 
        \hat u(\boldsymbol{x}_b; \boldsymbol{\theta}(n)) \\ 
        \mathscr{L}[\hat u](\boldsymbol{x}_r; \boldsymbol{\theta}(n)) 
    \end{bmatrix} 
    = \left( \boldsymbol{I} - e^{-\boldsymbol{K}(0)n} \right) 
    \begin{bmatrix} 
        g(\boldsymbol{x}_b) \\ \\
        f(\boldsymbol{x}_r) 
    \end{bmatrix}, \\
    \text{equivalently,} \quad
    &\begin{bmatrix} 
        \hat u(\boldsymbol{x}_b; \boldsymbol{\theta}(n)) \\ 
        \mathscr{L}[\hat u](\boldsymbol{x}_r; \boldsymbol{\theta}(n)) 
    \end{bmatrix} 
    - 
    \begin{bmatrix} 
        g(\boldsymbol{x}_b) \\ 
        f(\boldsymbol{x}_r) 
    \end{bmatrix} 
    = 
    - e^{-\boldsymbol{K}(0)n} 
    \begin{bmatrix} 
        g(\boldsymbol{x}_b) \\ 
        f(\boldsymbol{x}_r) 
    \end{bmatrix}.
\end{aligned}
\label{eq:ntk1}
\end{equation}\\

This is how the training error evolves with iterations. It can be further simplified using the semi-positive definiteness of NTK. We decompose $\boldsymbol K (0)$ as $\boldsymbol  K (0) = \boldsymbol Q ^T \boldsymbol \Lambda \boldsymbol Q$, where $\boldsymbol Q$ is an orthogonal matrix, and $\boldsymbol \Lambda$ is a diagonal matrix with eigenvalues $\lambda_i \geq 0$ as diagonal elements. With this, we rewrite \ref{eq:ntk1} as

\[\begin{bmatrix} \hat u(\boldsymbol{x}_b; \boldsymbol \theta(n)) \\ \mathscr{L}[\hat u](\boldsymbol{x}_r; \boldsymbol \theta(n)) \end{bmatrix} - \begin{bmatrix} g(\boldsymbol{x}_b) \\ f(\boldsymbol{x}_r) \end{bmatrix} = - \boldsymbol Q^T e^{- \boldsymbol \Lambda n} \boldsymbol Q \begin{bmatrix} g(\boldsymbol{x}_b) \\ f(\boldsymbol{x}_r) \end{bmatrix}.\]

This provides an estimate of the convergence of a PINN, which is governed by the magnitude of the eigenvalues of the NTK. Specifically, the larger the eigenvalue $\lambda_i$, the faster the $i$-th error component decays. Detailed mathematics of the above analysis can be found in \cite{MR4337814}. Later in the results section, we use these findings to establish a much faster convergence of WPINN over PINN for stiff problems.  

%%%%%%%%%%%%%%%%%%%%%%%%%%%%%%%%%%%%%%%%%%%%%%%%%%%%%%%%%%%%%%%%%%%%%%

\subsection{Wavelet-based physics-informed neural networks (W-PINNs)}\label{subsec_wavelets}

{The proposed W-PINN is designed as an integrated three-module architecture that decouples local feature extraction, multiscale solution representation, and  derivative evaluation. Rather than directly approximating the solution in physical space, W-PINN reformulates
physics-informed learning as a coefficient identification problem in a pre-computed wavelet basis. This design enables the neural network to focus on learning scale-aware coefficients, while deterministic wavelet operators handle reconstruction and differentiation.}

The W-PINN formulation begins by defining a family of wavelet basis functions, constructed as scaled and translated versions of a mother wavelet, which can be written as 
\begin{equation*}
    \Psi_{j,k}(x) = \sqrt{2^j}\psi(2^j x - k), \quad j,k \in \mathbb{Z}.
\end{equation*}
Here, $j$ denotes the scale (or resolution level) and $k$ is the translation parameter. For a compact domain $[a, b]$, the number of wavelet basis functions is determined by a predefined set of resolution levels $J = \{ J_1, J_2, \ldots, J_N \}.$ For a fixed $j \in J$, the translation indices $k$ span the interval $\left[ \left\lfloor (a-\gamma) \cdot 2^j \right\rfloor, \left\lceil (b+\gamma) \cdot 2^{j} \right\rceil \right],$ where $\gamma$ is a translation hyperparameter (of order of domain length), $\lfloor \cdot \rfloor$ and $\lceil \cdot \rceil$ denote floor and ceiling functions, respectively. This ensures sufficient coverage of the domain at each resolution level.

{After constructing a family of wavelets, the solution of the differential equation is represented in wavelet space, such that the learning task is shifted from point-wise solution approximation in the physical domain to coefficient learning across multiple resolutions:}

\begin{equation}\label{new_eq_2}
    \hat {u}(x) = \sum_{j=J_1}^{J_N}~\sum_{k=\lfloor (a-\gamma)\cdot 2^j \rfloor}^{\lceil (b+\gamma) \cdot 2^j \rceil} c_{j,k}\Psi_{j,k}(x) + \mathcal{B},
\end{equation}
where $c_{j,k}$ are trainable coefficients and $\mathcal{B}$ is a trainable bias to allow the representation of nonzero-mean solutions, given that the chosen wavelets are symmetric and inherently zero-mean. Extension of this formulation for higher-dimensional problems is straightforward, for instance, here is a formulation for a 2D problem:

\begin{equation}
    \begin{aligned}
        \Psi_{j_1,j_2,k_1,k_2}(x,y) &= \sqrt{2^{j_1} \cdot 2^{j_2}}\psi_X(2^{j_1} x - k_1)~\psi_Y(2^{j_2} y - k_2),\\
        \hat{u}(x,y) &= \sum_{j_1=J_{11}}^{J_{1N_1}}~ \sum_{j_2=J_{21}}^{J_{2N_2}} ~\sum_{k_1= \lfloor (a_1-\gamma) \cdot 2^{j_1} \rfloor}^{ \lceil (b_1+\gamma) \cdot 2^{j_1} \rceil} ~ \sum_{k_2= \lfloor (a_2-\gamma) \cdot 2^{j_2} \rfloor}^{\lceil (b_2+\gamma) \cdot 2^{j_2} \rceil} c_{j_1,j_2,k_1,k_2}\Psi_{j_1,j_2,k_1,k_2}(x,y) + \mathcal{B}.
    \end{aligned}
    \label{new_eq_4}
\end{equation}

{Here $\psi_X$ and $\psi_Y$ represent 1D wavelet functions in $x$ and $y$ directions, respectively.} This hierarchical construction of wavelet families at different resolutions and representing the solution in wavelet space enables capturing features at multiple scales, from broad, global behavior to fine, local details. In this study, we consider three mother wavelets, namely, the Gaussian, the Mexican Hat and the real Morlet. The mathematical expressions for the Mexican hat $(\psi^M(x))$, Gaussian $(\psi^G(x))$ and real Morlet $(\psi^{\mathcal{M}}(x))$can be written as 
\begin{equation*}
    \psi^M(x) = (1-x^2)e^{-\frac{x^2}{2}},~~~~
    \psi^G(x) = -xe^{-\frac{x^2}{2}},~~~~
    \psi^{\mathcal{M}}(x) = \cos(x)e^{-\frac{x^2}{2}}.
\end{equation*}

\begin{figure}[t!]
    \centering
    \includegraphics[width=0.99\linewidth]{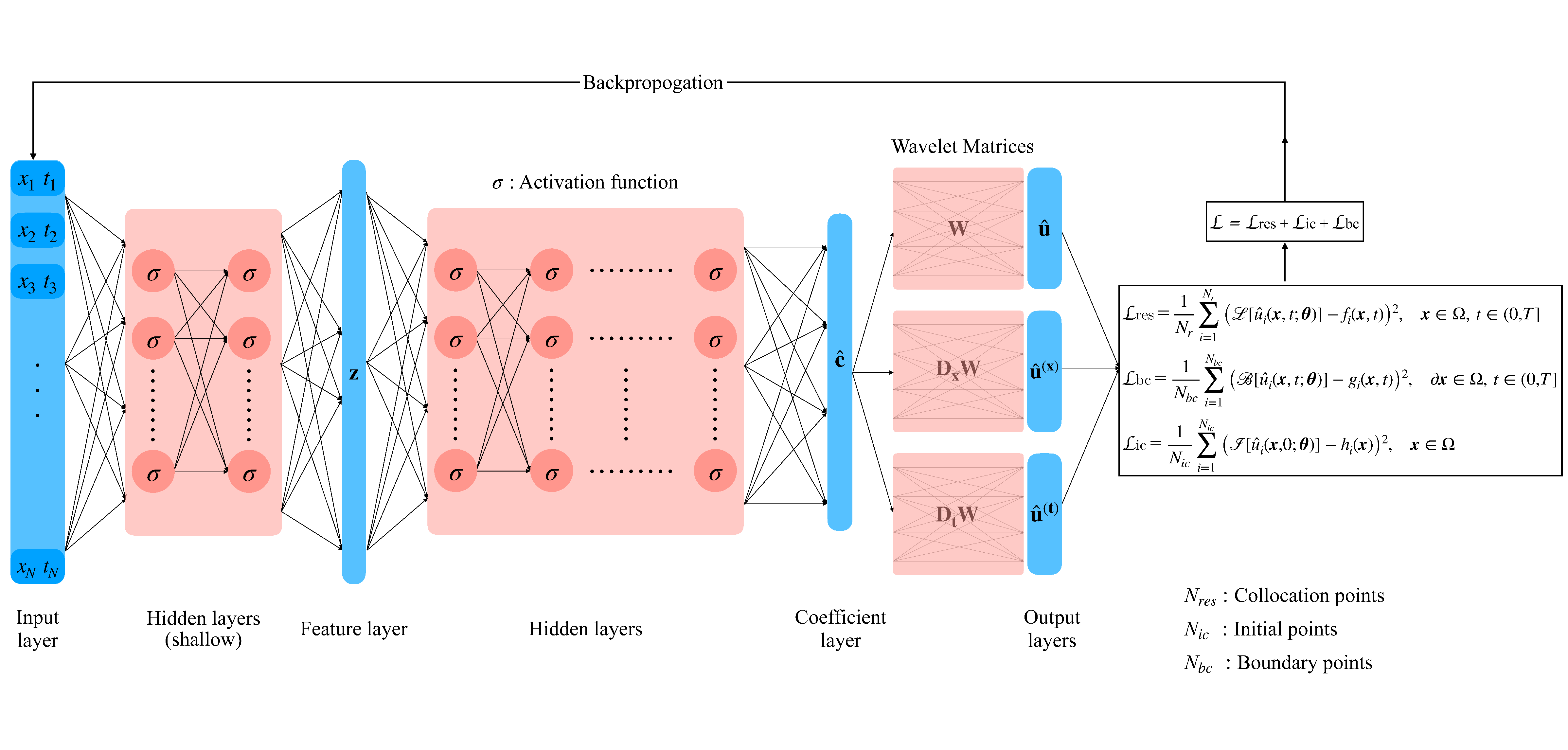}
    \caption{A schematic W-PINN architecture.}
    \label{fig:1}
\end{figure}

{The W-PINN architecture comprises three components, each serving a distinct and complementary role in addressing multiscale representation, optimization stiffness, and computational efficiency:

\begin{enumerate}
    \item \textbf{Pointwise feature mapping}: A shallow fully-connected network that maps
    spatial-temporal coordinates to a latent representation, capturing local nonlinear interactions
    without attempting to directly approximate oscillatory or singular solution components.
    
    \item \textbf{Global coefficient predictor}: A deep fully-connected network that predicts
    wavelet coefficients $c_{j,k}$, effectively learning the solution in wavelet space where
    multiscale features are naturally separated across resolutions.
    
    \item \textbf{Non-trainable inverse mapping}: A fixed layer that reconstructs the solution
    in physical space using predicted coefficients and pre-computed wavelet basis matrices,
    enabling exact and efficient derivative evaluation.

\end{enumerate}
}

The architecture of W-PINN for a two-dimensional problem is illustrated in Figure \ref{fig:1}. In general, for a $d$-dimensional problem, the network architecture comprises an input layer with $d$ neurons corresponding to the spatial-temporal dimensions. These inputs are processed through a shallow network that maps each collocation point to a corresponding entry in the feature layer. For one-dimensional problems, this mapping can be simplified by directly using the input layer as the feature layer. Now, the feature layer passes through a fully connected network to get coefficients. To further enhance model performance, these coefficients can be separately fine-tuned through hyperparameter optimization. The final solution and its derivatives are then computed by combining these optimized coefficients with the pre-calculated wavelet matrices. A trainable bias is also added to the network to incorporate $\mathcal{B}$. 
 
 For a given set of collocation points $\{x_0, x_1, \cdots, x_N\}$ and resolution set $J = [J_1,J_2,\cdots,J_n]$, using the domain $[0,1]$ and $\gamma = 0$, the wavelet matrices are constructed as follows: 
\begin{small}
\begin{equation*}
\mathbf{W} =
\begin{pmatrix}
\psi_{J_1,0}(x_0) & \cdots & \psi_{J_1,2^{J_1}}(x_0) & \cdots & \psi_{J_n,2^{J_n}}(x_0)\\
\psi_{J_1,0}(x_1) & \cdots & \psi_{J_1,2^{J_1}}(x_1) & \cdots & \psi_{J_n,2^{J_n}}(x_1)\\
\vdots & \vdots & \vdots & \vdots & \vdots \\
\psi_{J_1,0}(x_N) & \cdots & \psi_{J_1,2^{J_1}}(x_N) & \cdots & \psi_{J_n,2^{J_n}}(x_N)\\
\end{pmatrix},
~
\mathbf{D_1W} =
\begin{pmatrix}
\psi'_{J_1,0}(x_0) & \cdots & \psi'_{J_1,2^{J_1}}(x_0) & \cdots & \psi'_{J_n,2^{J_n}}(x_0)\\
\psi'_{J_1,0}(x_1) & \cdots & \psi'_{J_1,2^{J_1}}(x_1) & \cdots & \psi'_{J_n,2^{J_n}}(x_1)\\
\vdots & \vdots & \vdots & \vdots & \vdots \\
\psi'_{J_1,0}(x_N) & \cdots & \psi'_{J_1,2^{J_1}}(x_N) & \cdots & \psi'_{J_n,2^{J_n}}(x_N)\\
\end{pmatrix},
\end{equation*}
\end{small}
where $\psi'(x)$ is single derivate of $\psi(x)$ and similarly $\mathbf{D_2W}$ is constructed by double derivative of $\psi(x)$. {For a given resolution set $J$, the number of wavelet basis functions at the resolution level $J_i$ scales as $O(2^{J_i})$. Consequently, the total number of columns in the wavelet matrix, $\mathbf{W}$, grows as $\sum_{j \in J} O(2^{j})$.} Once the approximate solution and its derivatives are obtained, the loss function is computed as \ref{eq:loss}.

This wavelet-based formulation eliminates the need for automatic differentiation (AD) in computing PDE residuals, which not only makes training faster by reducing the complexity of the computational graph but also avoids potential numerical instabilities that commonly arise in AD during training, particularly in problems exhibiting high gradients or singularities. The core principle behind W-PINN's efficacy is that abrupt and multiscale features in the physical space become smoother in the wavelet domain, which improves conditioning of the optimization problem, leading to faster convergence and greater robustness in multiscale regimes.

%%%%%%%%%%%%%%%%%%%%%%%%%%%%%%%%%%%%%%%%%%%%%%
\section{Results and discussions}\label{Results}

In this section, we present the results obtained using the proposed method over a handful of examples and compare them with the performance of conventional PINNs and state-of-the-art methods. For a fair comparison, the parameters of all different methods are kept consistent. These parameters are reported alongside the examples and further summarized in Appendix \ref{A1}. We utilize the Adam optimizer \cite{kingma2014adam}, which combines two stochastic gradient descent algorithms: gradient descent with momentum and root mean squared propagation (RMSProp). RMSprop, also known as adaptive gradient descent, adapts the learning rate with iterations. The Adam optimizer is memory efficient and is well-suited for problems with high-dimensional parameter spaces.

The trainable parameters of networks are initialized using Xavier's technique or Glorot initialization \cite{glorot2010understanding}, which, in contrast to random initialization, initializes parameters in a suitable range to achieve faster convergence during training. During backpropagation, it maintains gradients within a reasonable range (not too low or high), thereby preventing the algorithm from getting stuck in poor local minima. Besides this, it can also reduce the need for extensive hyperparameter tuning, particularly the learning rate. We sampled the training set by Sobol sequencing throughout the domain, as we need efficient exploration of the entire input space for a single training set. For evaluation purposes, we employ the relative $L_2$ metric over uniform samples across the domain. 
\begin{equation}
    \text{Relative}~ L_2 ~\text{error} = \sqrt{\frac{\sum_{i=1}^M (u_i - \hat{u}_i)^2}{\sum_{i=1}^M u_i^2}},
\end{equation}
where $M$ represents the total number of testing samples, $u_i$ is the exact solution and $\hat{u}_i$ is the predicted one for the $i^{th}$ sample. In cases where the solution is not available analytically, we consider the exact solution as the numerical solution obtained using an in-house solver. 

For implementation purposes, PyTorch 2.4.1 with CUDA 12.1 is used. Training is done on the NVIDIA RTX A6000, which has 48GB of GPU memory. As the performance of a neural network is highly sensitive to the initialization of its parameters, we perform each experiment five times and report the relative $L_2$-error and average training time based on these five random runs. The source code of the proposed W-PINN method in this work is available on GitHub at {\url{https://github.com/himanshup21/W-PINN.git}}.

\begin{figure}[t!]
    \centering
    \includegraphics[width=0.8\linewidth]{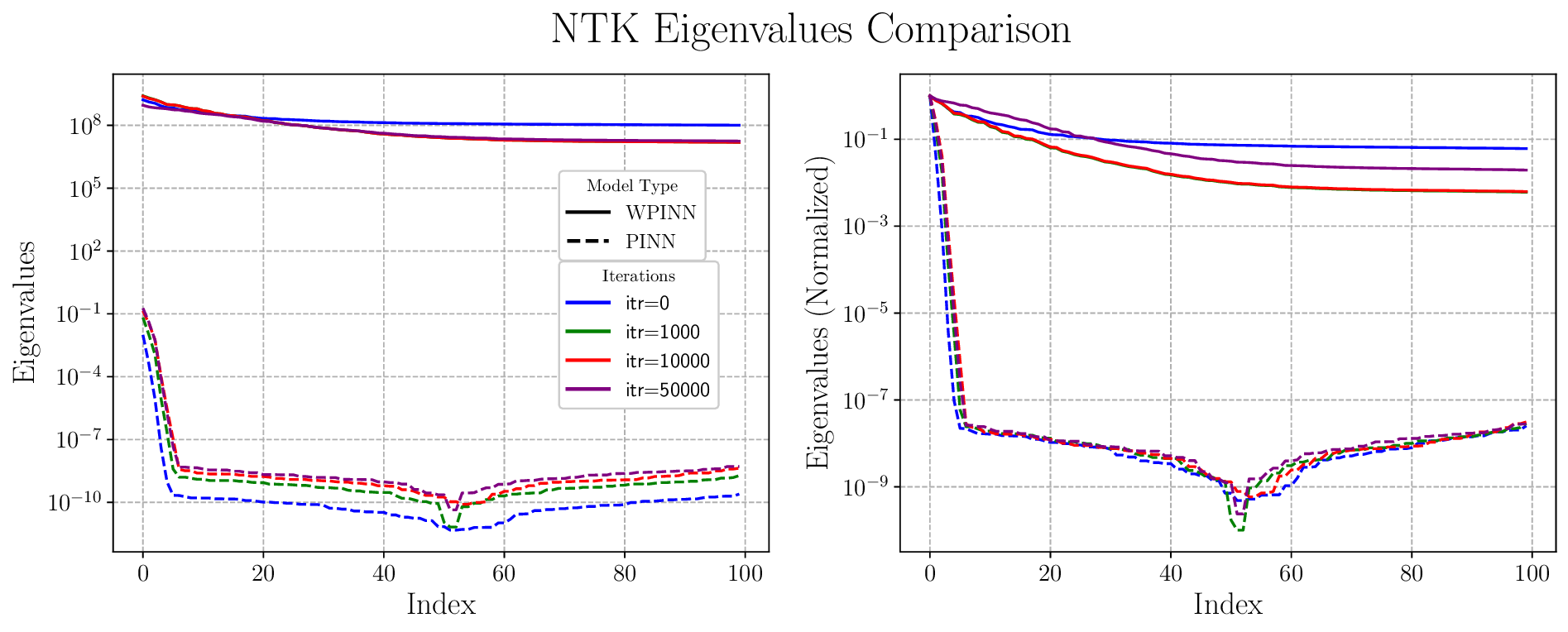}
    \caption{{{Comparison of NTK eigenvalues of PINN(dashed line) and W-PINN (solid line) for Example \ref{eg:1} with $\epsilon=2^{-7}$ at various iterations. Eigenvalues are sorted in descending order and plotted against the respective index.}}}
    \label{fig:NTK}
\end{figure}

\begin{table}[b!]
\begin{minipage}{.5\linewidth}
\caption{Parameters used for solving Example \ref{eg:1}.} 
\centering
		\begin{tabular}{l l}
			
			\hline
   &\\
			Parameters &Value\\
			\hline 
   &\\
   Translation hyperparameter $\gamma$ & ~1.0\\
			Set of resolutions ($J_M/J_G/J_{\mathcal{M}}$)&~[0,8]/[0,10]/[0,9]\\
			Number of hidden layers & ~6
			\\
			Neurons per layer & ~50 \\
   %Learning rate & ~$10^{-4}$ \\
   Number of collocation points & ~$10^{3}$  \\
   Maximum number of iterations & ~$5\times 10^{4}$  \\
      &\\
			\hline 
		\end{tabular}\label{tab:1}
  \end{minipage}
  \hspace{0.4cm}
  \begin{minipage}{.4\linewidth}
The\caption{{Average training time (in seconds) of PINN and W-PINN with network depth for Example \ref{eg:1}. } }
\centering
	{	\begin{tabular}{l l l l}
			
			\hline
   &&\\
			Depth &PINN&W-PINN&Speed-up\\
			\hline 
  & &\\
			~~2 & ~20.6 &~~~7.3 &~~~2.8x \\
			~~4 & ~27.2 &~~~6.2 &~~~4.4x \\
            ~~8 & ~47.6 &~~~8.3 &~~~5.7x \\
            ~~16 & ~82.6&~~~12.4 &~~~6.6x \\
            ~~32 & ~153.5 &~~~22.8 &~~~6.7x\\
            ~~64 & ~288.4 &~~~38.7 &~~~7.4x\\
     & &\\
			\hline 
		\end{tabular}}\label{tab:2}
  \end{minipage}
\end{table}

\begin{example}\label{eg:1}
    Consider the following one-dimensional linear advection-diffusion equations \cite{MR1721985}:
    \begin{equation}
        \begin{cases}
            \epsilon \dfrac{d^2 u}{d x^2}+(1+\epsilon)\dfrac{d u}{d x}+u=0,~x\in(0,1), \\[6pt]
        u(0)=0,~u(1)=1.
        \end{cases}
    \end{equation}
Here, $0<\epsilon\ll 1$. The exact solution to the considered problem is 
\begin{equation*}
    u(x)=\dfrac{\exp(-x)-\exp(-\frac{x}{\epsilon})}{\exp(-1)-\exp(-\frac{1}{\epsilon})}.
\end{equation*}
\end{example}
Such models, called Friedrich's boundary layer models, are often used to show how difficult it is to model viscous flow boundary layers \cite{MR4511357}. It is impossible for PINNs to capture the singularity that happens when $\epsilon \rightarrow 0$.

\begin{table}[t!]
\begin{center}
\caption{ {Performance comparison of various methods for solving Example \ref{eg:1}.}}
{
\begin{tabular}{cccc}
\hline
$\epsilon$ & Method & $L_2$-error & \begin{tabular}[c]{@{}c@{}}Average\\Training Time\end{tabular} \\
\hline
&&&\\
& PINN &  7.9e-05& 1m 58s \\
& Wavelet Activation \cite{uddin2023wavelets}& 3.9e-05  & 52s \\
& {PIRBN} \cite{PIRBN} &  {9.7e-05} & {2m 21s} \\
& {Gabor PINN}\cite{GaborPINN} &  {2.1e-04} & {3m 15s} \\
$\epsilon = 2^{-4}$ & W-PINN (Gaussian) & 2.5e-05 &  23s\\
& W-PINN (Mexican hat) &  8.1e-05& 48s \\
& {W-PINN (Morlet)} &  {7.2e-05}& {1m 33s} \\
\\
& PINN &  0.83& - \\
& Wavelet Activation \cite{uddin2023wavelets}&  0.85 & - \\
& {PIRBN}\cite{PIRBN} &  {0.81} & {-} \\
& {Gabor PINN} \cite{GaborPINN}&  {8.7e-2} & {4m 17s} \\
$\epsilon = 2^{-7}$ & W-PINN (Gaussian) & 5.3e-04  & 18s \\
& W-PINN (Mexican hat) & 9.2e-04  & 23s \\
& {W-PINN (Morlet)} &  {8.3e-04}& {1m 41s} \\
\\
& PINN & 0.84 & - \\
& Wavelet Activation \cite{uddin2023wavelets}& 0.85 & - \\
& {PIRBN} \cite{PIRBN}&  {0.85} & {-} \\
& {Gabor PINN} \cite{GaborPINN}&  {0.85} & {-} \\
$\epsilon = 2^{-10}$ & W-PINN (Gaussian) & 3.1e-03 & 38s \\
& W-PINN (Mexican hat) &  4.2e-03 & 42s \\
& {W-PINN (Morlet)} &  {3.1e-03}& {1m 55s} \\
&&&\\
\hline
\end{tabular}}
\label{tab:3_new}
\end{center}
\end{table}

\begin{figure}[h!]
    \centering
    \includegraphics[width=1.0\linewidth]{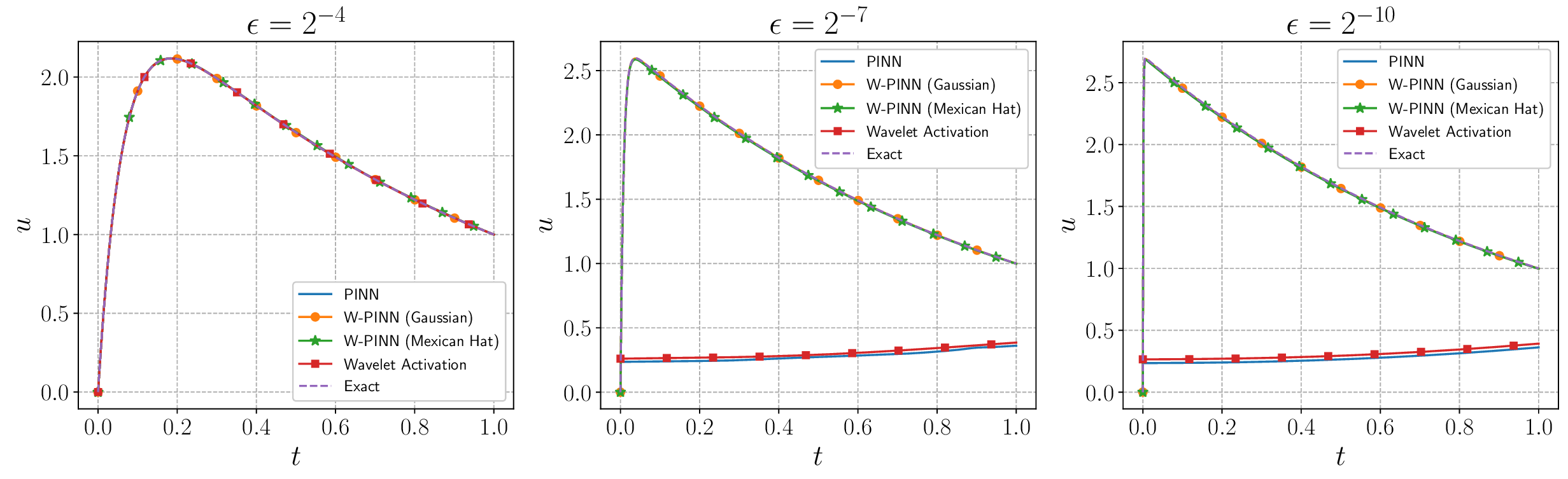}
    \caption{Comparison of solutions obtained by PINN, with wavelet activation, and W-PINN methods for Example \ref{eg:1}. From left to right: with $\epsilon = 2^{-4}$, $\epsilon = 2^{-7}$, and $\epsilon = 2^{-10}$.}
    \label{fig:4}
\end{figure}

The parameters of the W-PINN for the Example \ref{eg:1} are listed in Table \ref{tab:1}. Table \ref{tab:2} compares the training time of automatic differentiation based PINN and W-PINN. The experiments were conducted with $\epsilon = 2^{-4}$ and each test was executed for $10^4$ iterations. The results demonstrate that conventional PINN exhibits a steeper growth in training time with an increase in network depth, requiring up to $7.4$ times longer than W-PINN at depth $64$. This substantial difference in computational overhead can be attributed to W-PINN's elimination of automatic differentiation calculations, which become increasingly expensive in conventional PINNs as network depth increases. Moreover, Figure~\ref{fig:NTK} compares the Neural Tangent Kernel (NTK) eigenvalue spectra of W-PINN and standard PINN for $\epsilon = 2^{-7}$ across multiple training iterations. The top hundred NTK eigenvalues are sorted in descending order and plotted against their respective indices. {The left panel shows the NTK eigenvalues, while the right panel presents the same spectra normalized by their maximum eigenvalue to highlight relative decay behavior. The NTK eigenvalues of standard PINNs exhibit a rapid decay across modes, implying that higher-frequency components are associated with very small eigenvalues and therefore converge extremely slowly. In contrast, the spectrum of W-PINN decays significantly more slowly, indicating improved coupling to higher-frequency modes. This flatter spectral profile suggests that W-PINN can learn multiscale features more efficiently, leading to faster and more stable convergence.}\\

The comparison of various methods has been presented in Table \ref{tab:3_new}, which provides the relative $L_2$-error and average training time of the W-PINN, conventional PINN, PINN with gaussian-wavelet activation function \cite{uddin2023wavelets}, {PIRBN \cite{PIRBN}, and Gabor PINN \cite{GaborPINN}}. This table shows that the W-PINN approximates the solution with a much smaller $L_2$-error for lower values of $\epsilon$. Similar observations are drawn from Figure \ref{fig:4}, which demonstrates that when $\epsilon=2^{-4}$, there is no such sharp singularity near the origin, and all methods approximate the solution well. However, as $\epsilon$ decreases, the exact solution of the problem has a strong singularity near the origin. Then, both PINN and PINN with a wavelet activation function are unable to capture the behavior of the solution accurately. In contrast, W-PINN effectively resolves the singularity, with the predicted solution closely following the exact solution.

 \begin{example}\label{eg:2}
    Consider the following highly singularly perturbed non-linear problem \cite{MR3264586}:
    \begin{equation}
        \begin{cases}
            \epsilon \dfrac{d^2 u}{d t^2}+(3+t)\dfrac{d u}{d t}+u^2-\sin(u)=f,~t\in(0,1], \\[6pt]
        u(0)=1,~u^{\prime}(0)=1/\epsilon,
        \end{cases}
    \end{equation}
    where f is chosen such that $u(t)=2-\exp(-t/\epsilon)+t^2$ is the exact solution.
\end{example}

\begin{table}[b!]
\caption{Parameters used for solving Example \ref{eg:2}.} 
\centering
		\begin{tabular}{l l}
			
			\hline
   &\\
			Parameters &Value\\
			\hline 
   &\\
   
            Translation hyperparameter $\gamma$ & ~1.0\\
			Set of resolutions ($J_M/J_G/J_{\mathcal{M}}$)& ~[0,9]/[0,10]/[0,10]\\
			Number of hidden layers & ~6
			\\
			Neurons per layer & ~50 \\
  Number of collocation points & ~$10^{4}$  \\
   Maximum number of iterations & ~$10^{5}$  \\
   
      &\\
			\hline 
		\end{tabular}\label{tab:3}
\end{table}

\begin{table}[h!]
\begin{center}
\caption{ {Performance comparison of various methods for solving Example \ref{eg:2}. } }
{\begin{tabular}{cccc}
\hline
$\epsilon$ & Method & $L_2$-error & \begin{tabular}[c]{@{}c@{}}Average\\Training Time\end{tabular} \\
\hline
&&&\\
& PINN &  1.2e-04& 31s \\
& Wavelet Activation \cite{uddin2023wavelets}& 2.1e-04  & 58s \\
& {PIRBN}\cite{PIRBN} &  {4.7e-05} & {85s} \\
& {Gabor PINN}\cite{GaborPINN} &  {3.3e-04} & {55s} \\
$\epsilon = 2^{-4}$ & W-PINN (Gaussian) & 2.3e-05 & 48s \\
& W-PINN (Mexican hat) &  1.8e-04 & 51s \\
& {W-PINN (Morlet)} &  {4.1e-04} & { 1m 53s} \\

\\
& PINN &  8.5e-02&  3m 33s\\
& Wavelet Activation \cite{uddin2023wavelets}&  2.2e-02 &  3m 14s\\
& {PIRBN} \cite{PIRBN}&  {6.6e-04} & {2m 5s} \\
& {Gabor PINN}\cite{GaborPINN} &  {7.8e-03} & { 3m 17s} \\
$\epsilon = 2^{-7}$ & W-PINN (Gaussian) & 8.3e-05  & 2m 21s \\
& W-PINN (Mexican hat) &3.7e-04  & 2m 24s \\
& {W-PINN (Morlet)} &  {7.4e-05} & {2m 31s} \\
\\
& PINN & 1.4 & - \\
& Wavelet Activation \cite{uddin2023wavelets}& 0.34 & - \\
& {PIRBN} \cite{PIRBN}&  {1.6e-02} & {3m 13s} \\
& {Gabor PINN} \cite{GaborPINN}&  {9.1e-02} & { 3m 42s} \\
$\epsilon = 2^{-10}$ & W-PINN (Gaussian) & 6.6e-05  & 2m 27s \\
& W-PINN (Mexican hat) &  4.5e-04 & 2m 48s \\
& {W-PINN (Morlet)} &  {5.3e-05} & {2m 13s} \\
&&&\\
\hline
\end{tabular}}\label{tab:5_new}
\end{center}
\end{table}

\begin{figure}[b!]
    \centering
    \includegraphics[width=1.0\linewidth]{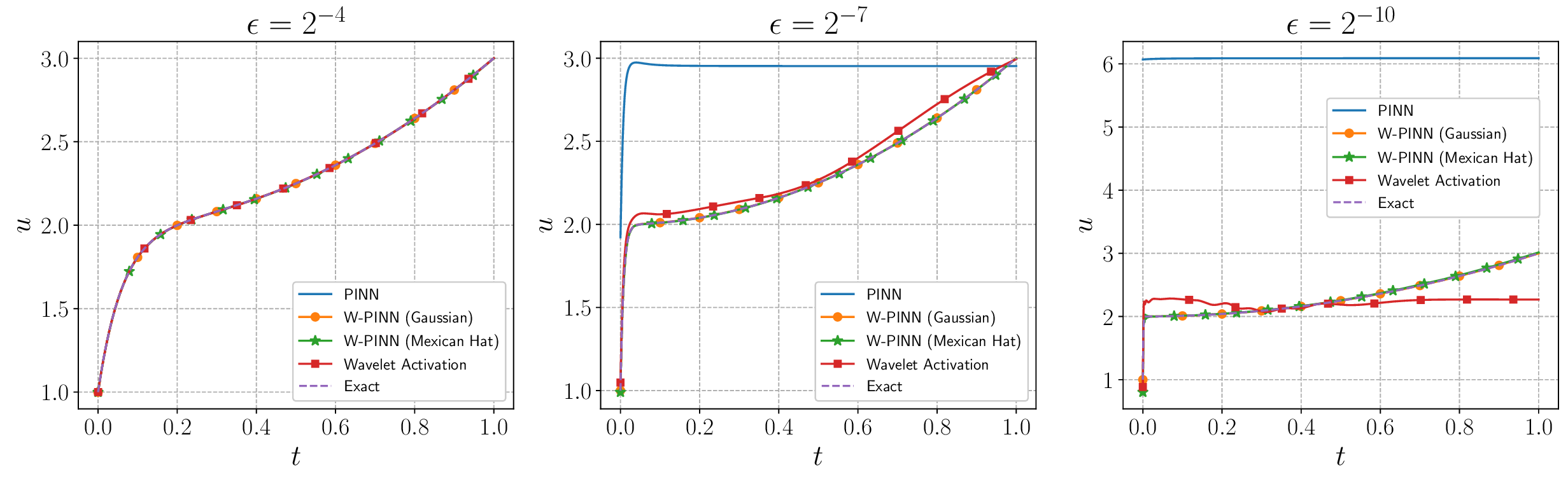}
    \caption{Comparison of solutions obtained by PINN, PINN with wavelet activation, and W-PINN methods for Example \ref{eg:2}. From left to right: with $\epsilon = 2^{-4}$, $\epsilon = 2^{-7}$, and $\epsilon = 2^{-10}$.}
    \label{fig:5}
\end{figure}

This example corresponds to a highly singularly perturbed non-linear equation with a known solution. For Example \ref{eg:2}, the network parameters are given in Table \ref{tab:3}. 

{As $\epsilon$ decreases, the initial layer becomes more pronounced, leading to increased stiffness and multiscale complexity, which becomes challenging to resolve. This behavior is reflected in Table~\ref{tab:5_new}, where the proposed W-PINN consistently maintains strong approximation capability across all values of $\epsilon$ for different wavelet choices, including Gaussian, Mexican hat, and Morlet wavelets. In contrast, the accuracy of competing approaches such as PINN with a Gaussian wavelet as activation function \cite{uddin2023wavelets}, PIRBN \cite{PIRBN}, and Gabor PINN \cite{GaborPINN} deteriorates as $\epsilon$ decreases, while W-PINN remains stable. These results highlight the effectiveness of wavelet-based representations in capturing fine-scale features induced by small $\epsilon$.}
 
Figure \ref{fig:5} demonstrates similar observations, for $\epsilon=2^{-10}$, both conventional PINN and PINN with the Gaussian wavelet as activation function \cite{uddin2023wavelets} are unable to detect the singularity satisfactorily, whereas W-PINN effectively handles the singularity.

\begin{figure}[t!]
  \begin{minipage}{0.47\textwidth}
    \centering
    \includegraphics[width=\textwidth]{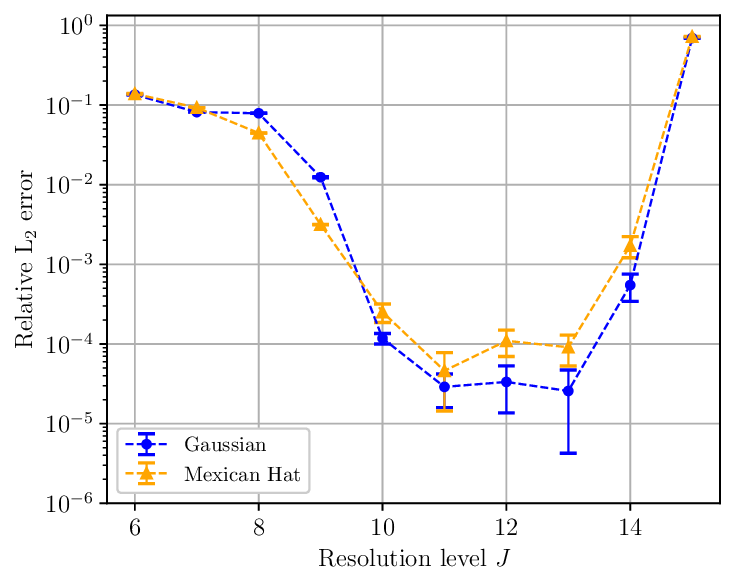}  
    \caption{{A plot for relative $L_2$-error with different wavelet resolution levels $([0, J])$ for Gaussian (blue) and Mexican-Hat (orange) wavelets.}}
    \label{fig:res}
  \end{minipage}
  \hfill
  \begin{minipage}{0.48\textwidth}
    \centering
    \includegraphics[width=\textwidth]{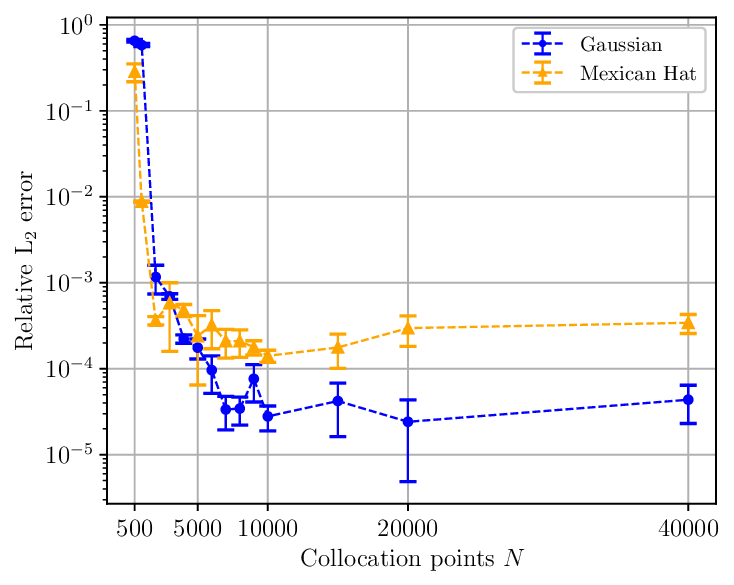}  
    \caption{{A plot for relative $L_2$-error variation with number of collocation points $(N)$ for Gaussian (blue) and Mexican-Hat (orange) wavelets.}}
    \label{fig:coll}
  \end{minipage}
\end{figure}

In addition, we investigate the relationship between the hyperparameters of W-PINN and its performance for this example. Figure \ref{fig:res} illustrates the relationship between the resolution level ($[0,J]$) and the relative $L_2$-error. Here, we employ a network with $8$ hidden layers, each containing $100$ neurons, and fix $10,000$ collocation points. It shows that W-PINN gives optimal results for a specific resolution range (not too high nor too low). Lower resolution is incapable of capturing sharp singularities, and a large $J$ makes the loss function too complex to be optimized. {The relatively small error bars demonstrate that W-PINN is numerically stable and well-conditioned with respect to random initialization.}

\begin{figure}[t!]
		\centering
		\begin{subfigure}{.47\textwidth}
			\centering
			\includegraphics[width=1\linewidth]{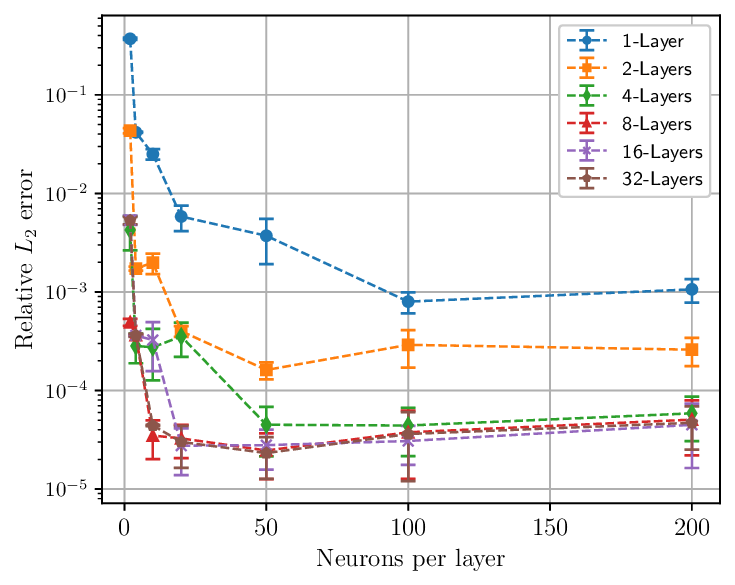}
			{$(a)\hspace{0.1cm}$ Gaussian wavelet}
		\end{subfigure}%
  \hspace{0.4cm}
		\begin{subfigure}{.48\textwidth}
			\centering
\includegraphics[width=1\linewidth]{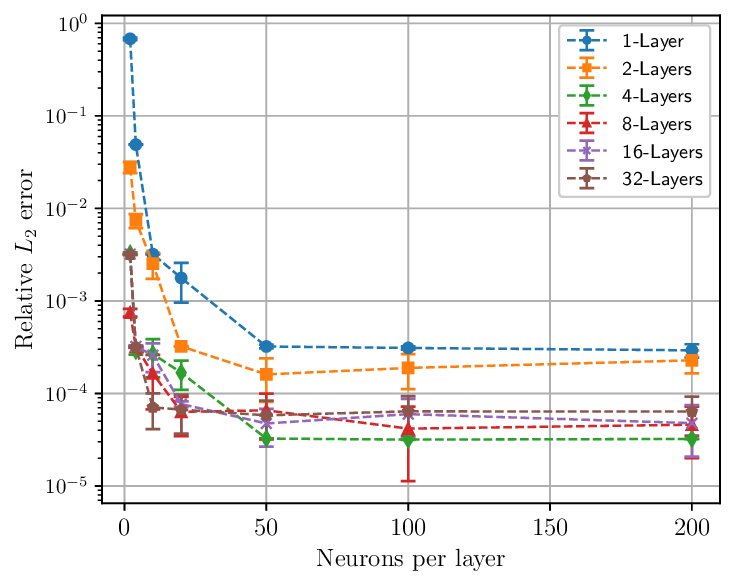}
			{$(b)\hspace{0.1cm}$ Mexican wavelet}
		\end{subfigure}
		\caption{{Impact of number of hidden layers and number of neurons per layer on performance of W-PINN using Gaussian wavelet(left) and Mexican-Hat wavelet(right).}}\label{various_hidden_layers}
	\end{figure}

Figure \ref{fig:coll} presents the effect of the number of collocation points on performance. We consider the identical network as previously used, with the Gaussian resolutions ($J_G = [0,12]$) and the Mexican hat resolutions ($J_M = [0,11]$). It is evident that the relative $L_2$-error decreases as the number of collocation points increases. However, the error decrease is not significant beyond a certain threshold. After a certain number of collocation points, additional points do not come with any new information about the behavior of the problem under consideration. 

Figure \ref{various_hidden_layers} presents experiments on the trainable part of W-PINN. We employ $10,000$ collocation points for both Gaussian and Mexican hat, with resolutions of $[0,12]$ and $[0,11]$, respectively. It implies that a shallow network is incapable of learning the behaviors of the problem, regardless of the number of neurons per layer. However, a network with a significant number of layers and sufficient neurons per layer performs adequately. It is redundant to employ a large number of layers and neurons, which increases computational cost rather than accuracy. 

\begin{example}\label{eg:3}
    Consider the following singularly perturbed non-linear problem with Neumann boundary conditions 
    \begin{equation}
        \begin{cases}
            -\epsilon \dfrac{d^2 u}{d t^2}+u^5+3u-1=0,~t\in[0,1], \\[6pt]
        u^{\prime}(0)=\sin(0.5),~u^{\prime}(1)=\exp(-0.7).
        \end{cases}
    \end{equation}
\end{example}
This example corresponds to a singularly perturbed non-linear problem with Neumann boundary conditions. This problem possesses a boundary layer singularity at both ends. 
The exact solution to this problem is unknown, so we obtained a numerical solution using Scipy solver and treated it as an exact solution.  
\begin{table}[t!]
\caption{Parameters used for solving Example \ref{eg:3}.} 
\centering
		\begin{tabular}{l l}
			
			\hline
   &\\
   
			Parameters &Value\\
			\hline 
   &\\
            Translation parameter $\gamma$&~1.0\\ 
			Set of resolutions ($J_M/J_G$)&~[0,8]/[0,9]\\
			Number of hidden layers & ~8
			\\
			Neurons per layer & ~50 \\
 Number of collocation points & ~$10^{4}$  \\
  Maximum number of iterations & ~$10^{4}$  \\
   
      &\\
			\hline 
		\end{tabular}\label{tab:5}
\end{table}

\begin{table}[b!]
\begin{center}
\caption{ {Performance comparison of various methods for solving Example \ref{eg:3}} }
{\begin{tabular}{cccc}
\hline
$\epsilon$ & Method & $L_2$-error & \begin{tabular}[c]{@{}c@{}}Average\\Training Time\end{tabular} \\
\hline
&&&\\
& PINN &  8.6e-05& 2m 26s \\
& Wavelet Activation \cite{uddin2023wavelets}& 3.2e-04  & 4m 49s \\
$\epsilon = 2^{-4}$ & W-PINN (Gaussian) & 4.7e-05 & 28s \\
& W-PINN (Mexican hat) &  1.4e-04& 34s \\
\\
& PINN &  1.8e-03& 4m 41s \\
& Wavelet Activation \cite{uddin2023wavelets}&  4.1e-03 & 6m 13s \\
$\epsilon = 2^{-7}$ & W-PINN (Gaussian) & 8.9e-06  & 14s \\
& W-PINN (Mexican hat) &8.3e-05  & 21s \\
\\
& PINN & 5.3e-03& 7m 14s \\
& Wavelet Activation \cite{uddin2023wavelets}& 8.4e-03 & 12m 28s \\
$\epsilon = 2^{-10}$ & W-PINN (Gaussian) & 6.5e-06  & 13s \\
& W-PINN (Mexican hat) &  9.9e-05 & 28s \\
&&&\\
\hline
\end{tabular}}\label{tab:7_new}
\end{center}
\end{table}

\begin{figure}[t!]
    \centering
    \includegraphics[width=1.0\linewidth]{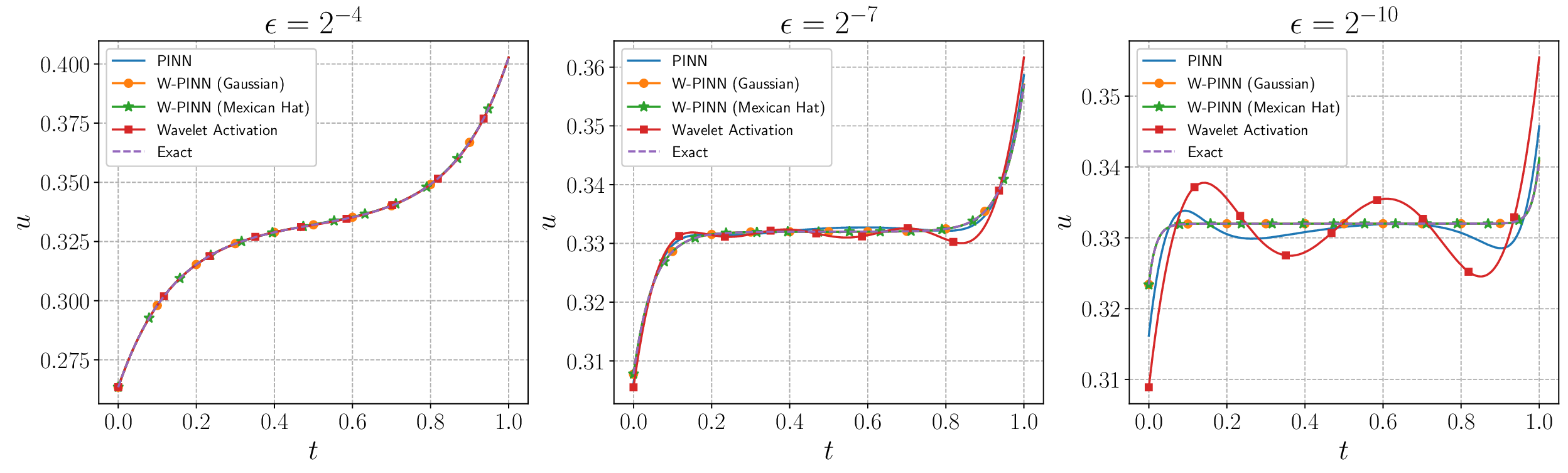}
    \caption{Comparison of solutions obtained by PINN, PINN with wavelet activation, and W-PINN methods for Example \ref{eg:3}. From left to right: with $\epsilon = 2^{-4}$, $\epsilon = 2^{-7}$, and $\epsilon = 2^{-10}$.}
    \label{fig:6}
\end{figure}
Table~\ref{tab:5} lists the network parameters used for Example~\ref{eg:3}. The corresponding numerical results are presented in Table~\ref{tab:7_new} and Figure~\ref{fig:6}. Specifically, the Table \ref{tab:7_new} compares the performance of W-PINN,  traditional PINN, and the PINN with a Gaussian wavelet as activation function \cite{uddin2023wavelets} for various $\epsilon$ in terms of relative $L_2$-error and average training time. This table shows that W-PINN takes much less training time and approximates the solution to a better extent.  Figure \ref{fig:6} demonstrates that as the singularity becomes more prominent, W-PINN effectively addresses the singularity to a greater degree at both ends. In contrast, other PINN methods couldn't resolve the singularity and introduced unnatural oscillations over the domain. 

\begin{example}\label{eg:4}
FitzHugh-Nagumo (FHN) model \cite{MR1237982}:
\end{example}
\noindent The FitzHugh-Nagumo (FHN) model is a simplified version of the Hodgkin-Huxley model, which simulates the dynamics of spiking neurons. It captures how a neuron generates electrical signals (spikes) in response to stimulation (an external current). The mathematical form of the FHN dynamical model is defined as follows: 
\begin{equation}\label{eq:7}
\begin{cases}
\dfrac{d v}{d t}-v+\dfrac{v^3}{3}+w-RI=0,\\[6pt]
\tau \dfrac{d w}{d t}-v+bw+a=0, \\
\end{cases}
\end{equation}
 where $v$ denotes the neuron's membrane voltage, $w$ is the recovery variable, reflecting the activation state of ion channels, $I$ represents the external current applied to the neuron, and $R$ is the resistance across the neuron. In addition, $a$ and $b$ are scaling parameters, and $\tau$ denotes the time constant for the recovery variable ($w$), which acts as a singularly perturbed parameter for this model. In particular, for $a=b=0$, the FHN model describes the Van der Pol oscillator, which describes self-sustaining oscillations in many systems, such as heartbeats, economies, and electronic circuits.

 \begin{table}[t!]
\caption{Parameters used for solving Example \ref{eg:4}.} 
\centering
		\begin{tabular}{l l}
			
			\hline
   &\\
			Parameters &Value\\
			\hline 
   &\\
            Translation parameter $\gamma$&~1.0\\ 
			Set of resolutions ($J_M/J_G$)&~[0,10]/[0,11]\\
			Number of hidden layers & ~10
			\\
			Neurons per layer & ~100 \\
  Number of collocation points & ~$10^{4}$  \\
   Number of iterations & ~$2\times 10^{4}$  \\

      &\\
			\hline 
		\end{tabular}\label{tab:9}
\end{table}

\begin{table}[t!]
\caption{{Performance comparison of various methods for solving Example \ref{eg:4} with $\tau=0.15$.}} 
\centering
	{	\begin{tabular}{l l l}
			
			\hline
 &  &\\
			Methods &~$L_2$-error &~~~~Average\\&~~~$(v/w)$&Training Time \\
			\hline 
  &&\\
			PINN &~$5.9e-04$ &~~~ 6m 38s\\
             &~$8.1e-02$ & \\ 
			Wavelet Activation &~$2.1e-03$ &~~~ 8m 22s\\
             &~$0.11$ & \\
            W-PINN (Gaussian) &~$9.2e-05$ &~~~ 1m 06s\\
             &~$5.7e-04$ & \\
            W-PINN (Mexican hat) &~$4.4e-03$ &~~~ 1m 36s\\
             &~$8.2e-03$ & \\

     & &\\
			\hline 
		\end{tabular}}\label{tab:9_new}
  \end{table}

\begin{figure}[t!]
    \centering
    \includegraphics[width=1.0\linewidth]{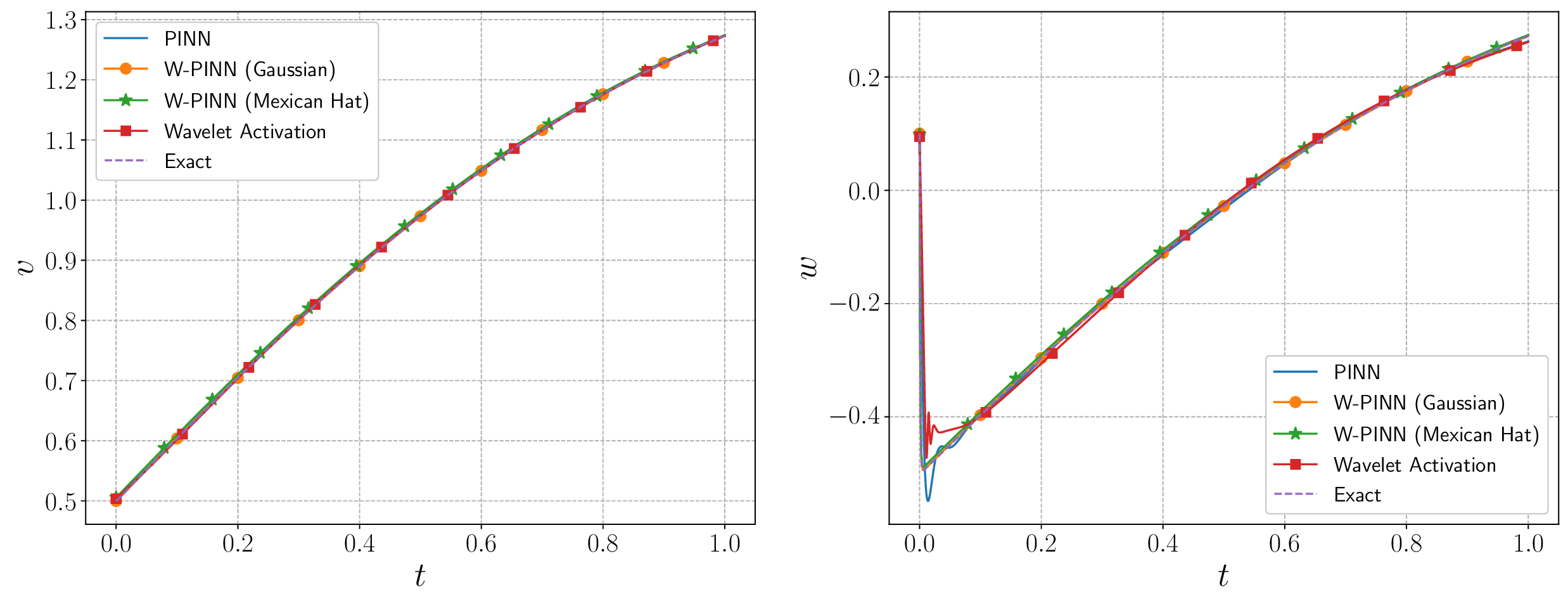}
    \caption{ Comparison of solution of FHN model \ref{eg:4} obtained by PINN, PINN with wavelet activation, and W-PINN methods with $\tau=2^{-10}$.}
    \label{fig:8}
\end{figure}

For the numerical computations, we set: $a=1,~b=1,~I=0.1,~R=1,$ and $\tau=2^{-10}$, $v(0)=0.5$ and $w(0) = 0.1$. The dynamics of the system are computed till $t=1$. Table \ref{tab:9} provides the parameters for this Example \ref{eg:4}. From Table \ref{tab:9_new}, it is evident that W-PINN gets a more accurate prediction and takes significantly less training time. Figure \ref{fig:8} compares the W-PINN approximations of $v$ and $w$ with PINN and PINN with wavelet activation. The figure illustrates that a strong initial layer is present in the solution profile of $w$, which both PINN and PINN with wavelet activation fail to capture and generate spurious oscillations near the singularity, whereas W-PINN effectively resolves the singularity.

\begin{example}\label{eg:5_new}
{A heat conduction problem with large gradients }\cite{lan2022dream}:
\end{example}
\noindent {The heat conduction problem investigates temperature flow when an intense heat source suddenly appears. It represents a practical challenge often seen in fusion applications, where researchers need to understand and model rapid heat transfer.}

\begin{table}[t!]
\caption{{Parameters used for solving Example \ref{eg:5_new}.}} 
\centering
	{	\begin{tabular}{l l}
			
			\hline
   &\\
			Parameters &Value\\
			\hline 
   &\\
   Translation hyperparameter $\gamma$ & ~0.2\\
	Set of resolutions $(J_x, J_t)$&~[-6,5], [-6,5]\\
			Number of hidden layers & ~6
			\\
			Neurons per layer & ~50 \\
      Number of collocation points & ~$ 10^{4}$  \\
      Number of boundary points & ~$ 10^{3}$  \\
      Number of initial points & ~$5\times10^2$ \\
    
      &\\
			\hline 
		\end{tabular}}\label{tab:111_new}
  \end{table}

\begin{table}[b!]
\caption{{Performance comparison of various methods for solving Example \ref{eg:5_new}.}} 
\centering
	{	\begin{tabular}{l l l}
			
			\hline
 &  &\\
			Methods &$L_2$-error &Average Training Time \\
			\hline 
  &&\\
			Conventional PINN&~$1.07$ \textpm 0.11 $\times 10^0$& ~-\\
			SA-PINN \cite{mcclenny2023self}& ~1.76 \textpm 0.35  $\times$ $10^{-3}$& ~66.75 min
			\\
            {PIRBN} \cite{PIRBN}&  {~7.73 \textpm 1.62 $\times 10^{-3}$} & { ~178.33 min} \\
            {Gabor PINN} \cite{GaborPINN} &  {~2.01 \textpm 2.14 $\times 10^{-3}$} & {~42.35 min} \\
			MMPINN-DNN \cite{wang2024practical} & ~$5.01$ \textpm 1.52 $\times 10^{-4}$ & ~11.02 min\\
   W-PINN (proposed method) & ~$2.56$ \textpm 1.3 $\times 10^{-4}$ & ~4.5 min\\

     & &\\
			\hline 
		\end{tabular}}\label{tab:10_new}
  \end{table}
  
{The following mathematical model includes a small positive constant $\epsilon$ that creates steep temperature gradients, making it a useful test case for our developed method:}
{ \begin{equation}
        \begin{cases}
             \dfrac{d u}{d t}=\dfrac{d^2 u}{d x^2}+f(x,t),~x\in(-1,1),~t\in[0,1], \\[6pt]
        u(x,0)=(1-x^2)\exp\left(\dfrac{1}{1+\epsilon}\right),~x\in(-1,1), \\[6pt] 
        u(-1,t)=0,~u(1,t)=0,~t\in (0, 1],
        \end{cases}
    \end{equation}}
  {  where $f$ is chosen such that $u(x,t) = (1-x^2)\exp\left(\dfrac{1}{(2t-1)^2+\epsilon}\right)$ is the exact solution.}

  The behavior of this model exhibits interesting characteristics depending on the value of the parameter $\epsilon$. 
 However, when $\epsilon$ becomes very small, the solution shows dramatic changes near $t=0.5$, exhibiting distinct multiscale 
characteristics. An important observation is the relationship between the supervised loss term and the residual term - while the boundary conditions ensure the supervised loss term remains minimal (as the boundary value is $0$ and the initial value stays below $\epsilon$), the residual term grows considerably as $\epsilon$ decreases. This is quantitatively demonstrated as, when $\epsilon=0.15$ the ratio of $\mathcal{L}_\text{bc} :\mathcal{L}_\text{ic} : \mathcal{L}_\text{res} = 1 : 10 : 10^7.$
Traditional PINN methods generally perform well for smooth problems, but face challenges 
when dealing with such large disparities between supervised and residual terms. Our proposed W-PINN can effectively handle such loss imbalances.

\begin{figure}[t!]
    \centering
    \includegraphics[width=1.0\linewidth]{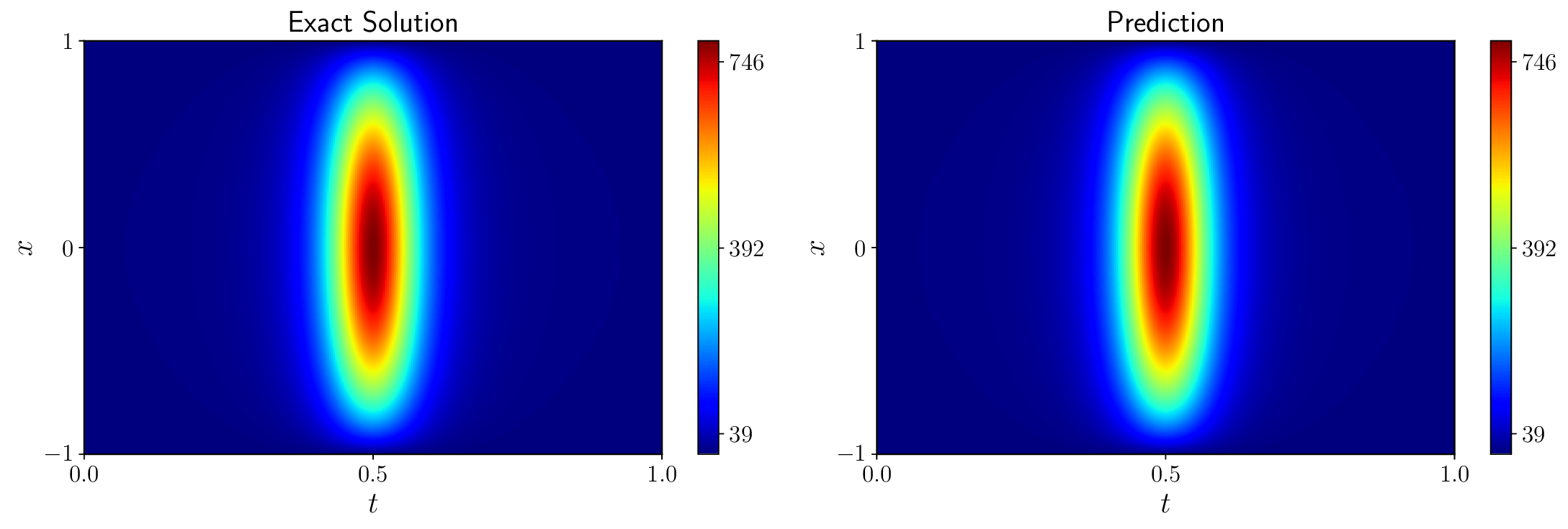}
    \caption{{The exact solution (left) and the prediction of W-PINN (right) with $\epsilon=0.15$ for high gradient heat conduction problem \ref{eg:5_new} .}}
    \label{fig:11_new}
\end{figure}

\begin{figure}[t!]
    \centering
    \includegraphics[width=.95\linewidth]{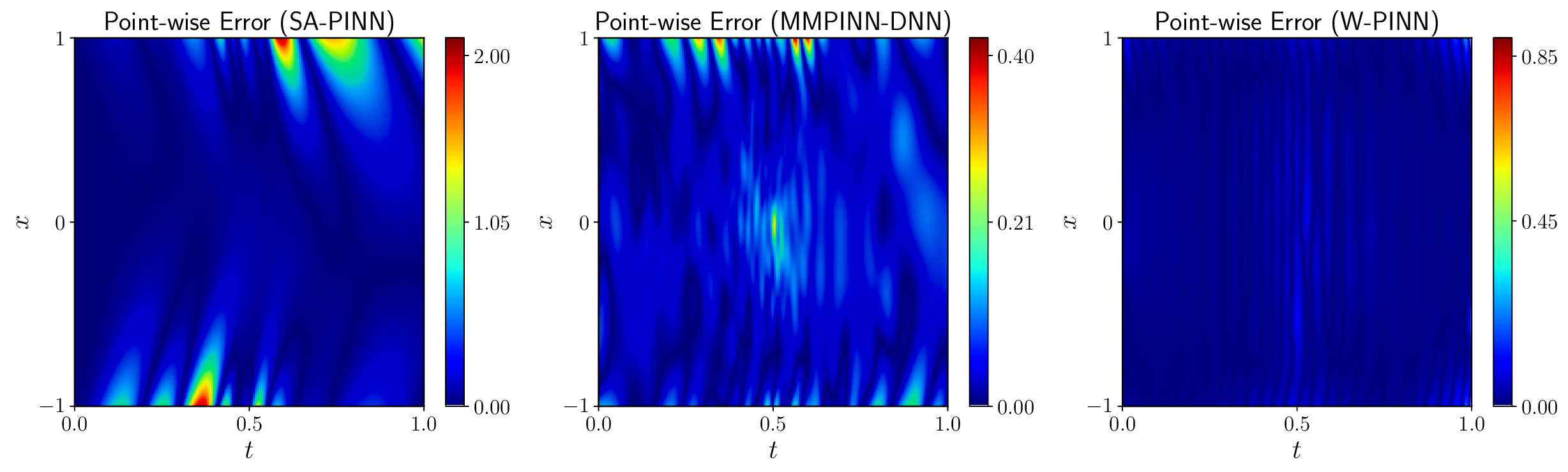}
      \caption{{Point-wise absolute error for Example \ref{eg:5_new}. From left to right: SA-PINN, MMPINN-DNN, and W-PINN.}}
    \label{fig:9_new}
\end{figure}

Table \ref{tab:111_new} represents the parameters of the network for this Example \ref{eg:5_new}. Further, Table~\ref{tab:10_new} presents a comparative analysis of different methods for solving Example~\ref{eg:5_new} with $\epsilon = 0.15$, evaluated in terms of relative $L_2$-error and average training time. The conventional PINN fails to produce an accurate approximation for this problem. The self-adaptive PINN (SA-PINN)~\cite{mcclenny2023self} significantly improves accuracy, however, this improvement is accompanied by a substantial increase in training time. {Methods such as PIRBN \cite{PIRBN} and Gabor PINN \cite{GaborPINN} also achieve improved accuracy compared to the conventional PINN, but they require considerably higher training time. The multi-magnitude PINN (MMPINN-DNN)~\cite{wang2024practical} attains better accuracy with moderate training time. In contrast, the proposed W-PINN provides the most favorable balance between accuracy and efficiency, achieving the lowest $L_2$-error among all compared methods while requiring significantly less training time. These results underscore the effectiveness of the wavelet-based formulation in efficiently capturing multiscale features.}

\begin{figure}[t!]
    \centering
    \includegraphics[width=0.9\linewidth]{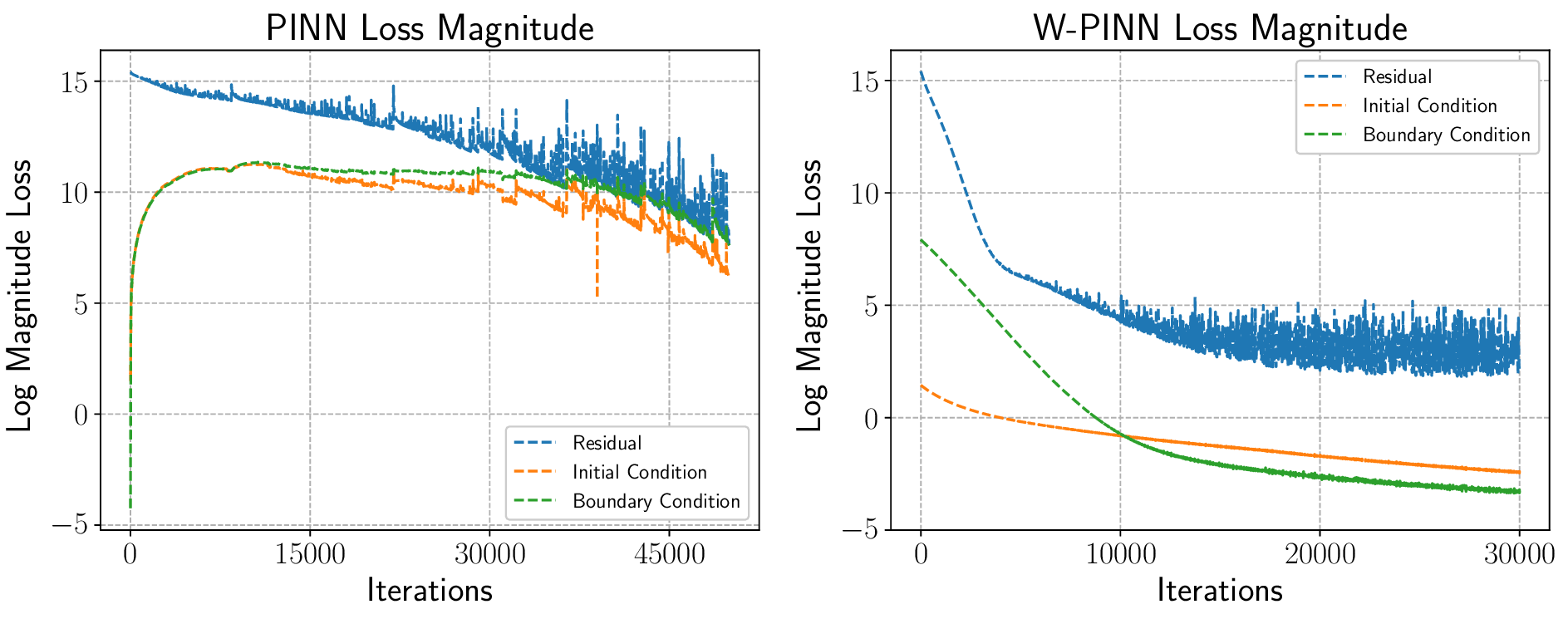}
    \caption{{Loss curve of PINN (left) and W-PINN (right) for Example \ref{eg:5_new} with $\epsilon=0.15$.}}
    \label{fig:10_new}
\end{figure}

{Figure \ref{fig:11_new} demonstrates the similarity between the exact solution and the W-PINN's prediction for Example \ref{eg:5_new}. Figure \ref{fig:9_new} shows point-wise absolute error comparisons across three approaches: SA-PINN, MMPINN-DNN, and W-PINN. This error plot shows that errors for W-PINN throughout the domain remain consistently low except for extremely small regions near corners. Further, Figure \ref{fig:10_new} demonstrates the training progress of conventional PINN and W-PINN by showing their loss components over iterations. PINN shows unstable behavior with the boundary and initial condition losses initially increasing, suggesting potential training difficulties. In contrast, W-PINN exhibits a much more stable and efficient training pattern, where all three loss components (residual, initial condition, and boundary condition) consistently decrease from the start, reaching lower magnitudes in fewer iterations. This indicates that W-PINN achieves better optimization performance than the standard PINN approach. }

\begin{table}[t!]
\caption{{Parameters used for solving Example \ref{eg:6_new}.}} 
\centering
	{	\begin{tabular}{l l}
			
			\hline
   &\\
			Parameters &Value\\
			\hline 
   &\\
   Translation hyperparameter $\gamma$ &~0.2\\
	Set of resolutions $(J_x, J_y)$&~[-4,5], [-4,5]\\
			Number of hidden layers & ~6
			\\
			Neurons per layer & ~50 \\
      Number of collocation points & ~$ 10^{4}$  \\
      Number of boundary points & ~$10^{3}$  \\

      &\\
			\hline 
		\end{tabular}}\label{tab:11_new}
  \end{table}
  
\begin{table}[t!]
\caption{{Performance comparison of various methods for solving Example \ref{eg:6_new}.}} 
\centering
	{	\begin{tabular}{l l l}
			
			\hline
 &  &\\
    Methods &$L_2$-error &Average Training Time \\
    \hline 
  &&\\
    Conventional PINN&~$4.93$ \textpm 1.56 $\times 10^{-2}$& ~23.18 min\\
    SA-PINN \cite{mcclenny2023self}& ~1.27\textpm 1.11$\times$ $10^{-2}$& ~152.88 min
    \\
    {PIRBN}\cite{PIRBN} &  {~8.82 \textpm 1.32 $\times 10^{-3}$} & { ~38.32 min} \\
            {Gabor PINN}\cite{GaborPINN} &  {~3.93 \textpm 2.07 $\times 10^{-3}$} & {~24.17 min} \\
    MMPINN-MFF\cite{wang2024practical} & ~$6.56$ \textpm 4.17 $\times 10^{-4}$ & ~22.75 min\\
  MMPINN-INN\cite{wang2024practical} & ~$2.13$ \textpm 0.43 $\times 10^{-4}$ & ~64.29 min\\
   W-PINN (proposed method) & ~$3.12$ \textpm 0.71$\times 10^{-4}$ & ~6.3 min\\

     & &\\
			\hline 
		\end{tabular}}\label{tab:12_new}
  \end{table}

\begin{figure}[t!]
    \centering
    \includegraphics[width=1.0\linewidth]{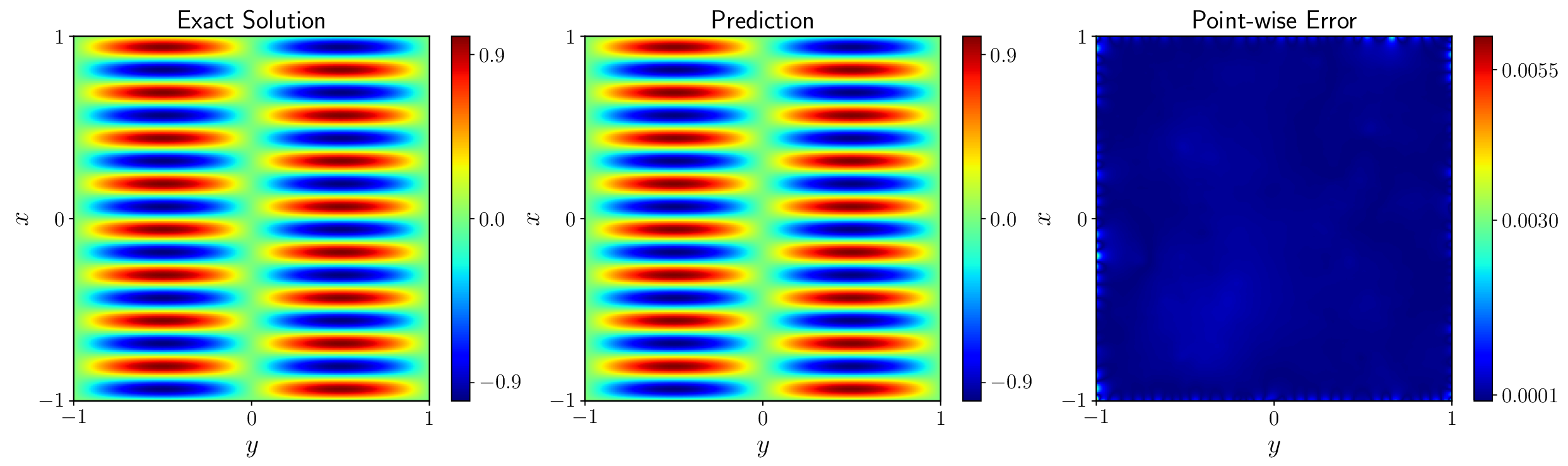}
    \caption{{From left to right: The exact solution, the predicted solution, and the point-wise absolute error using W-PINN for the Helmholtz equation \ref{eg:6_new}.} }
    \label{fig:12_new}
    \includegraphics[width=0.85\linewidth]{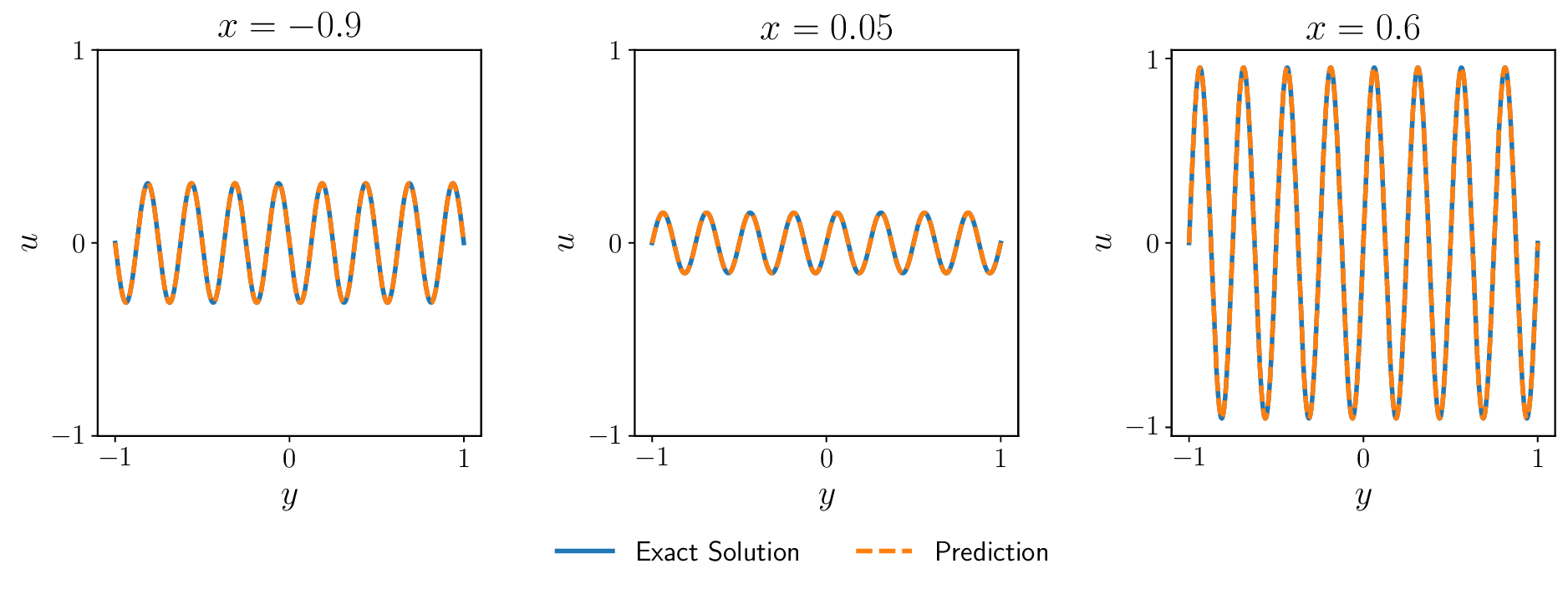}
    \caption{{Comparison of the exact solution (solid blue line) and the prediction (dashed orange line) using W-PINN at different spatial stamps for Example \ref{eg:6_new}.}}
    \label{fig:13_new}
\end{figure}

  \begin{example}\label{eg:6_new}
{Helmholtz equation with high-frequency:}
\end{example}
\noindent {The Helmholtz equation is a crucial elliptic partial differential equation that models electromagnetic wave behavior. It appears in wide applications in physics and engineering. The equation combines the Laplacian operator ($\Delta u$) with a wave number term ($c^2$), and is typically studied on bounded domains with appropriate boundary conditions. In the high-frequency regime, this equation becomes particularly challenging to solve numerically due to the oscillatory nature of its solutions.}
{
\begin{equation}
\begin{cases}
\Delta u(x,y) + c^2u(x,y) = f(x,y), & (x,y) \in \Omega = (-1,1) \times (-1,1), \\
u(x,y) = p(x,y), & (x,y) \in \partial\Omega.
\end{cases}
\end{equation}}
{$f$ and $p$ are determined in such a way $u(x,y) = \sin(b_1\pi x)\sin(b_2\pi y)$ is the exact solution.}

{Due to the significant initial imbalance between residual and supervised components, conventional PINN fails to provide accurate predictions as the optimization process becomes heavily skewed towards residual minimization \cite{wang2021understanding}.}

{To test the performance of W-PINN, we set the model parameters as $c = 1, b_1 = 1$, and $b_2 = 8$. Table \ref{tab:11_new} represents the parameters of the network for this Example \ref{eg:6_new}. The comparative analysis presented in Table \ref{tab:12_new} demonstrates that W-PINN achieves comparable or superior accuracy in terms of $L_2$-error relative to state-of-the-art methods in recent literature. Along with its significantly reduced training time, W-PINN is established as a particularly attractive framework. Figure \ref{fig:12_new} illustrates the remarkable agreement between the analytical solution and W-PINN predictions across the entire domain. In Figure \ref{fig:13_new}, cross-sectional comparisons at various x-coordinates further validate the method's accuracy, with predicted solutions indistinguishable from the exact solutions.}

\begin{example}\label{eg:7_new}
{Allen-Cahn reaction-diffusion equation}:
\end{example}
\noindent {The Allen-Cahn reaction-diffusion PDE is a widely used model in materials science, particularly for simulating phase separation processes in metallic alloys \cite{moelans2008introduction, kunselman2020semi}. The equation describes the evolution of an order parameter, $u$, which represents the phase state of the material. The PDE combines a diffusion term with a non-linear reaction term that drives the phase separation. The Allen-Cahn equation used in this study is defined as:}
{\begin{equation}
\begin{cases}
u_t - \epsilon u_{xx} + 5u^3 - 5u = 0, \quad x \in [-1, 1], \quad t \in [0, 1], \\
u(t, -1) = u(t, 1),\\
u_x(t,-1)=u_x(t,1),\\
u(x, 0) = x^2 \cos(\pi x),
\end{cases}
\end{equation}}
{here $\epsilon$ denotes a singularly perturbed parameter. For numerical computations, we choose $\epsilon=10^{-4}$. Moreover, the exact solution to this problem is unknown, so we obtained a numerical solution using Scipy solver and treated it as the exact solution. The Allen-Cahn equation is particularly challenging due to its rapid changes across space and time, making it difficult to model accurately. Additionally, its periodic boundary conditions provide an extra layer of complexity, making it an ideal benchmark for testing the model's capabilities. }

\begin{table}[t!]
\caption{{Parameters used for solving Example \ref{eg:7_new}.}} 
\centering
{		\begin{tabular}{l l}
			
			\hline
   &\\
			Parameters &Value\\
			\hline 
   &\\
            Translation parameter $\gamma$&~0.4\\ 
			Resolutions $(J_x, J_t)$&~[-5,6], [-5,5]\\
			Number of hidden layers & ~6
			\\
			Neurons per layer & ~100 \\
         Number of collocation points & ~$2\times 10^{4}$  \\
         Number of boundary points & ~$ 2\times 10^{3}$  \\
      Number of initial points & ~$ 10^{3}$  \\
      &\\
			\hline 
	\end{tabular}}\label{tab:14_new}
\end{table}

\begin{table}[t!]
 		\caption{{Performance comparison of various methods for solving Example \ref{eg:7_new} with $\epsilon = 10^{-4}$.}}
\centering
	{	\begin{tabular}{l l}
			
			\hline
   &\\
			Methods &$L_2$-error \\
			\hline 
   &\\
	Conventional PINN	& $0.96 \pm 0.06$ \\ 
 Time-adaptive approach \cite{wight2020solving} & $8.0\times 10^{-2}$$\pm$
$0.56\times 10^{-2}$\\
  SA-PINN \cite{mcclenny2023self} & $2.1 \times 10^{-2} \pm 1.21 \times 10^{-2}$\\ 
    W-PINN (proposed method)  &$4.8\times 10^{-2} \pm 0.6\times 10^{-2}$\\
     &\\
			\hline 
	\end{tabular}}\label{tab:15_new}
\end{table}

\begin{figure}[t!]
    \centering
    \includegraphics[width=1\linewidth]{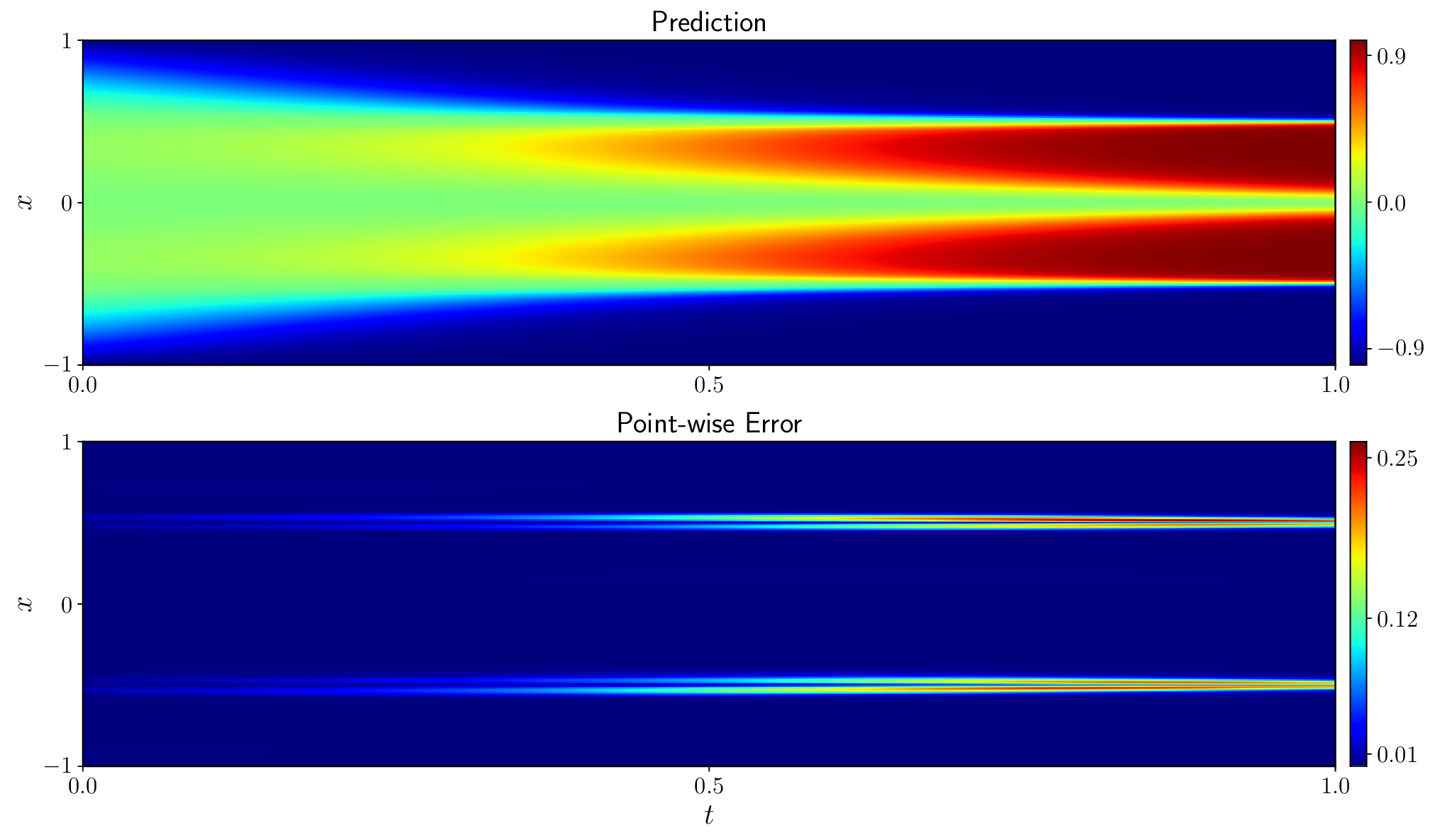}
\caption{{Top: Predicted solution of Allen-Cahn equation \ref{eg:7_new} via W-PINN. Bottom: point-wise absolute error distribution.}}\label{fig:14_new}
   \includegraphics[width=0.85\linewidth]{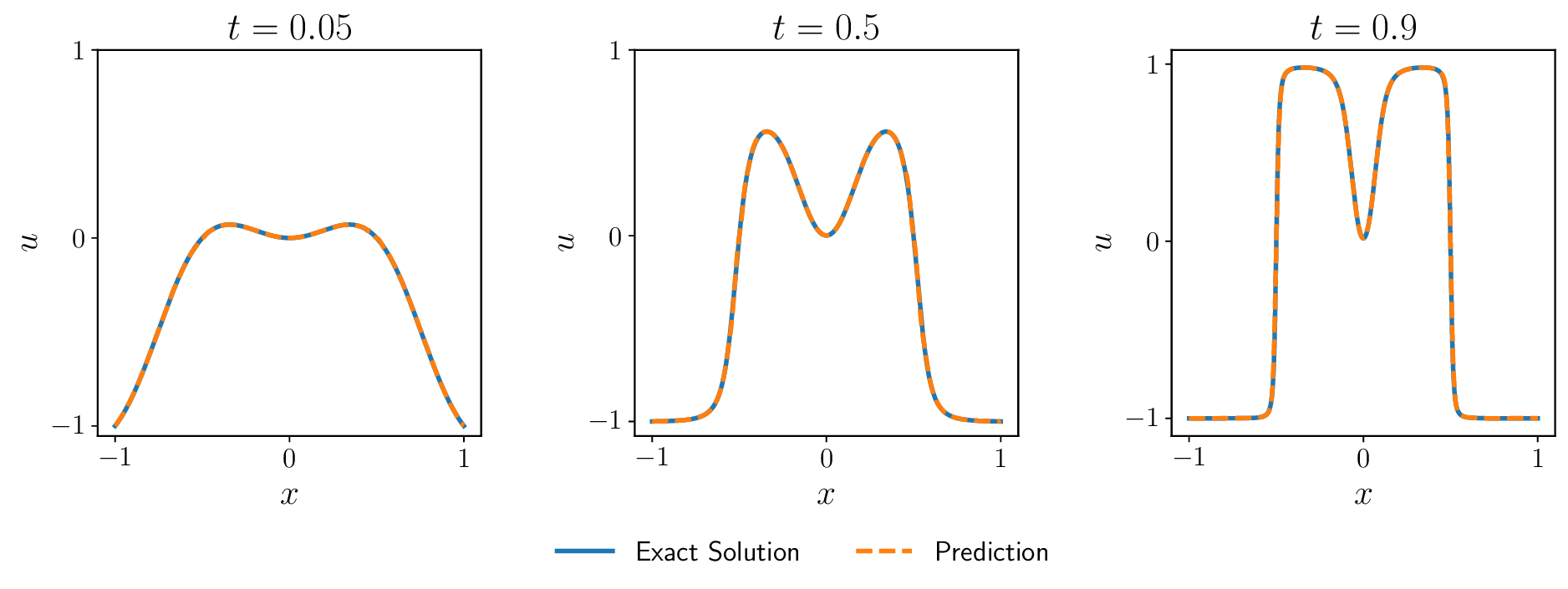}
    \caption{{Comparison of exact solution (solid blue line) and predicted solution (dashed orange line) using W-PINN at different time instances for Example \ref{eg:7_new}.}}
    \label{fig:15_new}
\end{figure}

{Table \ref{tab:14_new} provides the parameters used for this problem. Table \ref{tab:15_new} presents a comparative analysis of different PINN-based methods. The conventional PINN completely fails to capture the complex dynamics of the problem. While the time-adaptive method \cite{wight2020solving} shows some improvement, its accuracy remains unsatisfactory. In contrast, our proposed W-PINN achieves accuracy comparable to SA-PINN \cite{mcclenny2023self} while demonstrating computational efficiency with approximately twice the speed (~30 ms/iteration) of SA-PINN. Figure \ref{fig:14_new} illustrates the predicted dynamics using the W-PINN method, where the predominance of dark blue regions in the error distribution plot indicates consistently low prediction errors across the solution domain. Further, Figure \ref{fig:15_new} compares W-PINN prediction with the exact solution at three different time stamps, validating the model's ability to capture the temporal evolution of the system dynamics accurately. }

{
\begin{example}\label{eg:8_new}
{Maxwell's Equation}:
\end{example}
\noindent {Maxwell's equations represent the fundamental principles governing electromagnetic theory, describing the relationships between electric and magnetic fields and their interaction with matter. Their applications span numerous fields of engineering and physics. In telecommunications, these equations govern the propagation of electromagnetic waves through optical fibers and wireless channels. The medical industry relies heavily on Maxwell's equations for diagnostic imaging technologies, particularly in magnetic resonance imaging (MRI) \cite{brown2014magnetic}. In the aerospace sector, these equations are essential for radar system design and satellite communication systems. In their differential form, Maxwell's equations are expressed as:}

{\begin{equation}
\left\{
\begin{array}{ll}
    \nabla \times \boldsymbol{E}(t, \boldsymbol{x}) &= -\mu(\boldsymbol{x})\dfrac{\partial \boldsymbol{H}(t, \boldsymbol{x})}{\partial t},\\ [6pt]
    \nabla \times \boldsymbol{H}(t, \boldsymbol{x}) &= \varepsilon(\boldsymbol{x})\dfrac{\partial \boldsymbol{E}(t, \boldsymbol{x})}{\partial t},
\end{array}\right.
\end{equation}}

where $\boldsymbol{E}$ is the electric field vector, $\boldsymbol{H}$ is the magnetic field vector. The material properties are characterized by $\mu$, the magnetic permeability, which quantifies the medium's response to magnetic fields, and $\epsilon$, the electric permittivity, which describes the medium's capacity to store electrical energy. The operators $\nabla$ and $\times$ represent the gradient operator and cross product, respectively.

A significant challenge in solving these equations arises from their inherent rapid oscillations in both space and time. This characteristic poses particular difficulties for traditional PINNs, especially when dealing with heterogeneous media where $\mu$ and $\epsilon$ exhibit discontinuous behavior at media interfaces. Here, we present a solution of Maxwell's equation in both a homogeneous and a heterogeneous medium via W-PINN and compare them with the traditional PINN.

\begin{table}[t!]
\caption{{{Parameters used for solving Example \ref{eg:8_new}.}}} 
\centering
{		\begin{tabular}{l l}
			
			\hline
   &\\
			Parameters &Value\\
			\hline 
   &\\
			Translation hyperparameter $\gamma$&~0.5\\ 
            Resolutions $(J_x, J_t)$&~[-5,5], [-5,5]\\
			Number of hidden layers & ~8
			\\
			Neurons per layer & ~50 \\
         Number of collocation points & ~$10^{4}$  \\
         Number of boundary points & ~$10^3$  \\
      Number of initial points & ~$5\times10^2$ \\
      &\\
			\hline 
	\end{tabular}}\label{tab:16_new}
\end{table}

\begin{figure}[b!]
    \centering
    \includegraphics[width=0.9\linewidth]{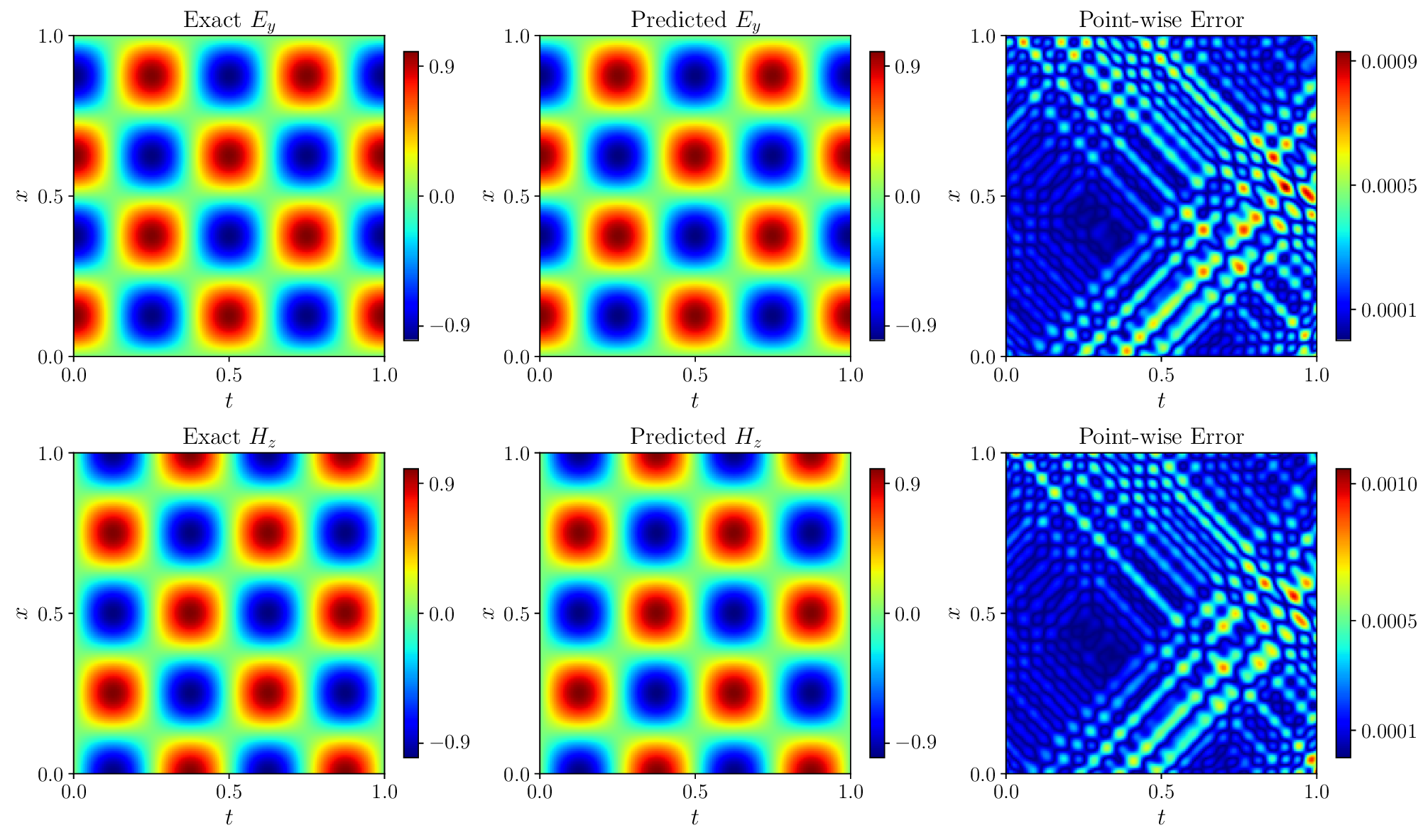}
\caption{{{Solution of Maxwell's equation \ref{homo} in homogeneous medium using W-PINN. Top: $E_y$ Exact, predicted, and corresponding point-wise absolute error.
Bottom: $H_z$ Exact, predicted, and corresponding point-wise absolute error.}}}\label{fig:16_new}
\end{figure}

\begin{figure}[b!]
    \centering
    \includegraphics[width=0.9\linewidth]{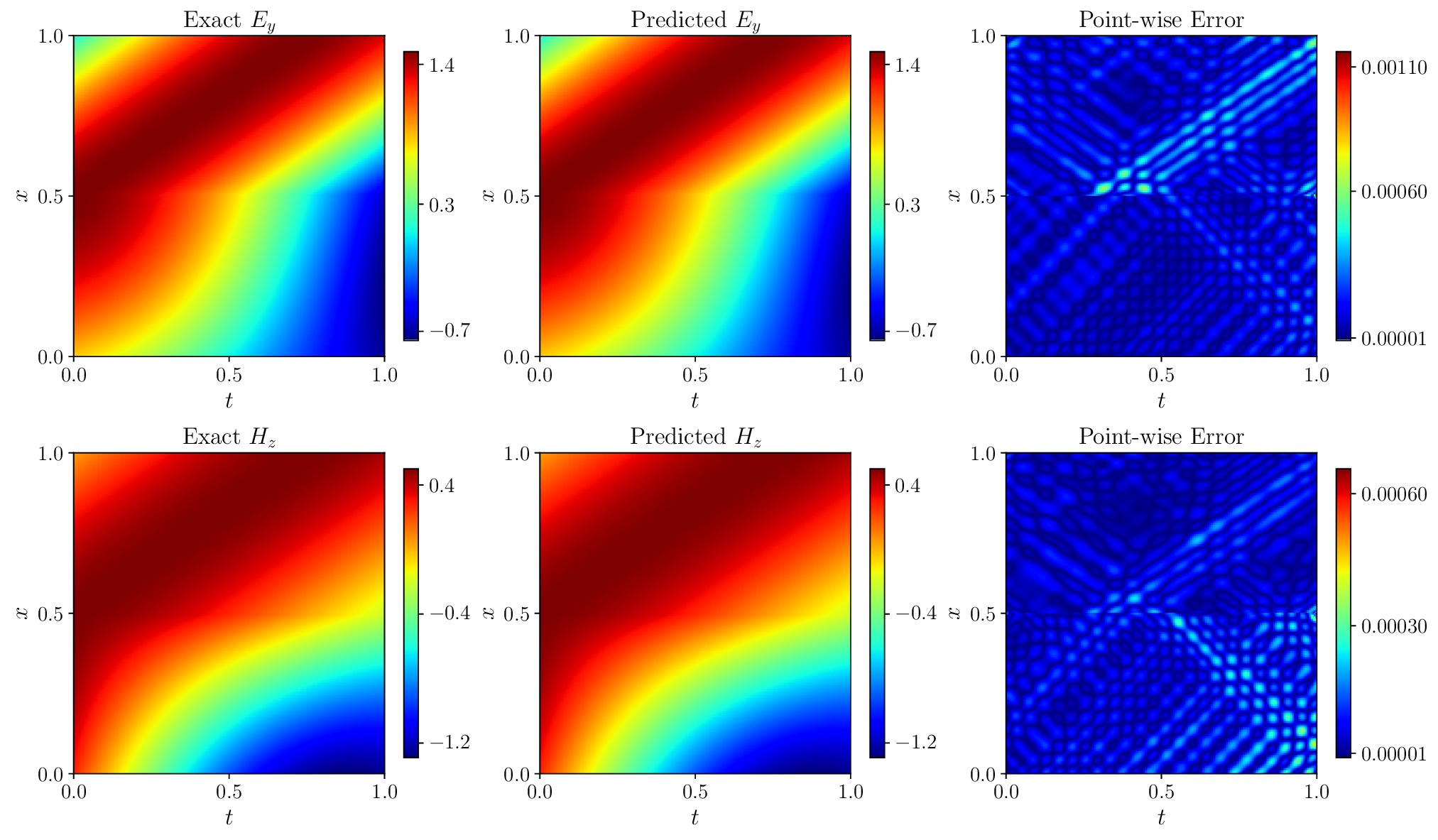}
\caption{{Solution of Maxwell's equation \ref{homo} in heterogeneous medium using W-PINN. Top: $E_y$ Exact, predicted, and corresponding point-wise absolute error.
Bottom: $H_z$ Exact, predicted, and corresponding point-wise absolute error.}}\label{fig:17_new}
\end{figure}

\begin{enumerate}
    \item[\textbf{A. } ] \textbf{Homogeneous media}
\end{enumerate}

We first investigate the performance of our proposed method by solving Maxwell's equations in a one-dimensional cavity model with homogeneous media. The governing equations are:

\begin{equation}\label{homo}
\frac{\partial E_y}{\partial t} = - \frac{1}{\epsilon} \frac{\partial H_z}{\partial x}, ~~~\frac{\partial H_z}{\partial t} = - \frac{1}{\mu} \frac{\partial E_y}{\partial x}, \quad x \in [0, 1], \quad t \in [0, 1].
\end{equation}

The system is subject to perfectly electric conductor (PEC) boundary conditions:

\begin{equation}
\begin{cases}
E_y(0,t) = E_y(1,t) = 0,\\
\dfrac{\partial H_z(x,t)}{\partial x}\bigg{|}_{x=0,1} = 0.
\end{cases}
\end{equation}

The ground truth for the example is given by:

\begin{equation}
\begin{cases}
E_y = \sin(n \pi x)\cos(\omega t),\\
H_z = -\cos(n \pi x)\sin(\omega t),
\end{cases}
\end{equation}
where $n=4$ is the order of cavity mode, cavity frequency $\omega = n \pi / l$ and $l$ is the length of medium.

Over five random runs, W-PINN achieved average relative $L_2$-errors of $5.47\pm 1.12\times10^{-4}$ and $5.51\pm2.01 \times 10^{-4}$ for $E_y$ and $H_z$, respectively. In contrast, traditional PINNs exhibited notably higher errors of $3.08\pm1.18 \times 10^{-3}$ and $5.18\pm1.93 \times 10^{-3}$. The computational efficiency of W-PINN is evident in its training time, {which averages 6.2 minutes over roughly $2\times10^4$ iterations on an Nvidia A6000 GPU.} The network architecture parameters are tabulated in Table \ref{tab:16_new}.
Figure \ref{fig:16_new} presents the W-PINN predictions for both electric and magnetic fields, alongside their corresponding point-wise errors within the cavity. The remarkably low point-wise error distribution validates the high accuracy of our approach in capturing the electromagnetic field behavior.

\begin{enumerate}
    \item[\textbf{B. } ] \textbf{Heterogeneous media}
\end{enumerate}

For this case, we chose Maxwell's equation \ref{homo} in a heterogeneous medium with the following analytical solution:  

\begin{equation}
E_y = 
\begin{cases}
    \cos(2t - 2x + 1) \\
    +0.5\cos(2t + 2x - 1), & \text{if } x \in \Omega_1, \\[8pt]
    1.5\cos(2t - 3x + 1.5), & \text{if } x \in \Omega_2,
\end{cases}
\end{equation}

\begin{equation}
H_z = 
\begin{cases}
    \cos(2t - 2x + 1) \\
    -0.5\cos(2t + 2x - 1), & \text{if } x \in \Omega_1, \\[8pt]
    0.5\cos(2t - 3x + 1.5), & \text{if } x \in \Omega_2,
\end{cases}
\end{equation}
where $\Omega_1 = [0,0.5]$ and $\Omega_2 = [0.5,1]$. Here $\mu=1, \epsilon=1$ in $\Omega_1$ and $\mu=4.5, \epsilon=0.5$ in $\Omega_2$. According to the electromagnetic interface condition: 

\begin{equation}
\begin{cases}
E_y(0.5,t)\big|_{x\in \Omega1} = E_y(0.5,t)\big|_{x\in \Omega2}, \\[6pt]
H_z(0.5,t)\big|_{x\in \Omega1} = H_z(0.5,t)\big|_{x\in \Omega2}.
\end{cases}
\end{equation}

In this challenging context of heterogeneous media, W-PINN demonstrates remarkable accuracy, achieving average relative $L_2$-errors of $2.41\pm 1.22\times10^{-4}$ and $2.81\pm1.51 \times 10^{-4}$ for electric ($E_y$) and magnetic ($H_z$) fields, respectively. These results represent a two-order-of-magnitude improvement over traditional PINNs, which exhibit substantially higher errors of $1.37\pm1.18 \times 10^{-2}$ and $2.07\pm0.51 \times 10^{-2}$. {W-PINN took an average of 16.1 minutes for about $3\times10^4$ iterations on the Nvidia A6000 GPU.} The network parameters are the same as in Table \ref{tab:16_new}.
Figure \ref{fig:17_new} illustrates the W-PINN predictions for both electric and magnetic fields, alongside their corresponding point-wise absolute errors. While the media interface presents the greatest computational challenge, W-PINN maintains impressive accuracy throughout the domain, successfully capturing the Electromagnetic field behavior across the media.

\begin{figure}[t!]
    \centering
    \includegraphics[width=0.5\linewidth]{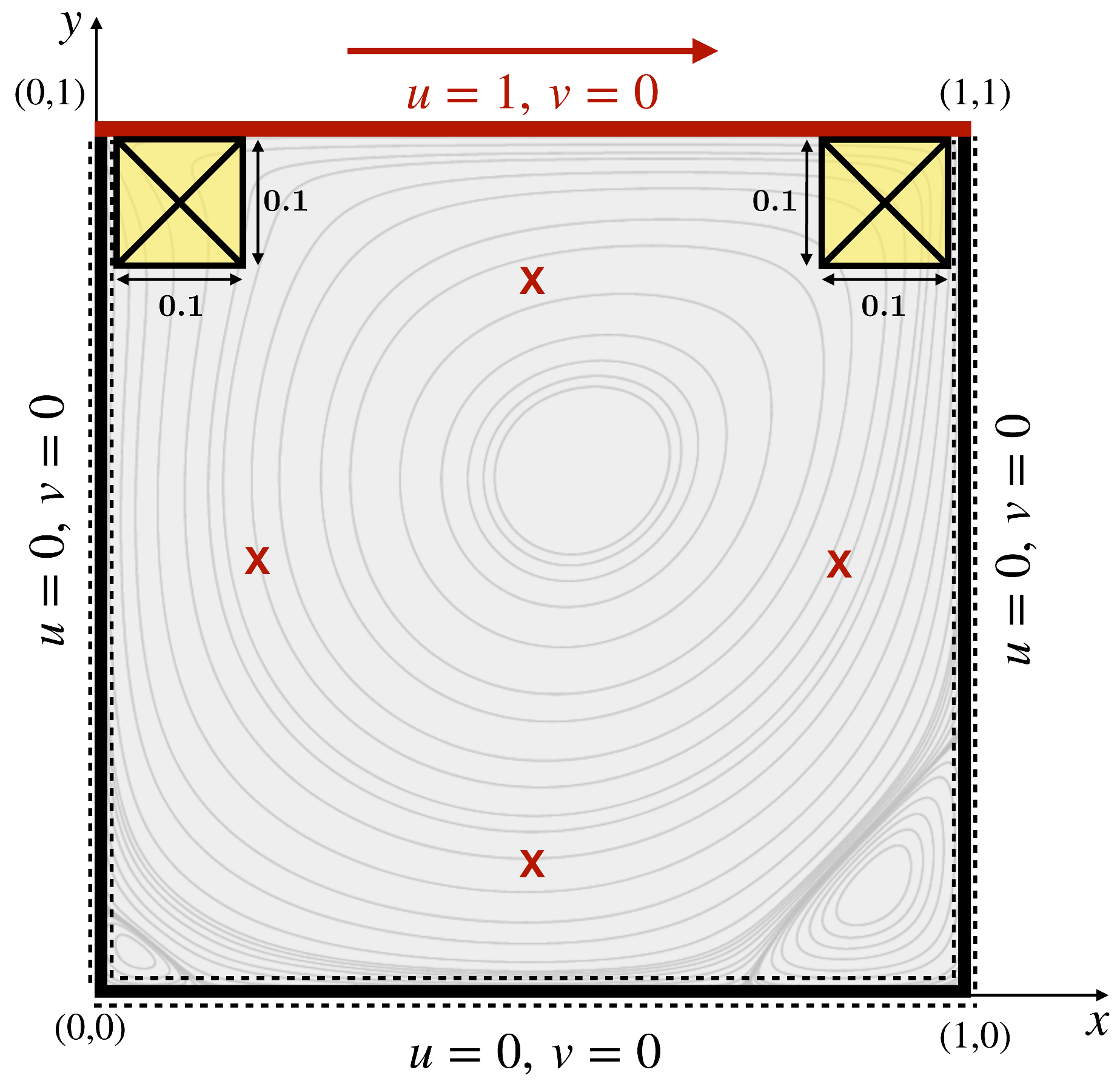}
\caption{{A lid-driven unit square cavity. For Re = 400: additional collocation points are sampled from the yellow crossed region, and red crosses indicate reference points.}}\label{fig:cavity}
\end{figure}

\begin{example}\label{eg:9_new}
{Lid-Driven Cavity flow}:
\end{example}
\noindent The steady state two-dimensional incompressible lid-driven cavity flow is a classic benchmark problem in computational fluid dynamics. The flow is governed by the incompressible Navier-Stokes equations, which present significant challenges due to their non-linear nature and the presence of pressure-velocity coupling. As illustrated in Figure \ref{fig:cavity}, the problem consists of a unit square cavity where the top wall moves with a uniform velocity of 1 unit while maintaining no-slip conditions on all other walls. The governing Navier-Stokes equations for this system are:

\begin{equation}\label{eg:_new9}
\left\{
\begin{array}{ll}
    \dfrac{\partial u}{\partial x} + \dfrac{\partial v}{\partial y} = 0,\\ [10pt]
    u\dfrac{\partial u}{\partial x} + v\dfrac{\partial u}{\partial y} + \dfrac{\partial p}{\partial x} - \dfrac{1}{\text{Re}}\left(\dfrac{\partial^2 u}{\partial x^2} + \dfrac{\partial^2 u}{\partial y^2}\right) = 0,\\ [10pt]
    u\dfrac{\partial v}{\partial x} + v\dfrac{\partial v}{\partial y} + \dfrac{\partial p}{\partial y} - \dfrac{1}{\text{Re}}\left(\dfrac{\partial^2 v}{\partial x^2} + \dfrac{\partial^2 v}{\partial y^2}\right) = 0,
\end{array}
\right.
\end{equation}
where $u$ and $v$ are the fluid velocities in $x$ and $y$ directions respectively, and $p$ is the fluid pressure. Re is the Reynolds number, which represents the ratio of inertial forces to viscous forces. We evaluate our W-PINN approach for two distinct flow regimes: Re = 100 and Re = 400, and compare the results with the solution obtained through COMSOL and also validate against the widely accepted benchmark solutions of Ghia et al.\cite{GHIA1982387}.\\

\begin{table}[t!]
\caption{{{Parameters used for solving Example \ref{eg:9_new}.}}} 
\centering
{		\begin{tabular}{l l}
            \hline
            &\\
            Parameters & Value \\
            \hline 
            &\\
            Translation hyperparameter $\gamma$ & ~0.5\\
            Resolutions $(J_x, J_y)$ & ~[-15,5], [-15,5] \\
            Number of hidden layers & ~12 \\
            Neurons per layer & ~100 \\
            Number of collocation points & ~$4\times 10^{4}$ \\
            Number of boundary points & ~$2\times10^3$ \\
            &\\
            \hline 
        \end{tabular}}
        \label{tab:17_new}
\end{table}

\begin{figure}[b!]
    \centering
    \includegraphics[width=0.9\linewidth]{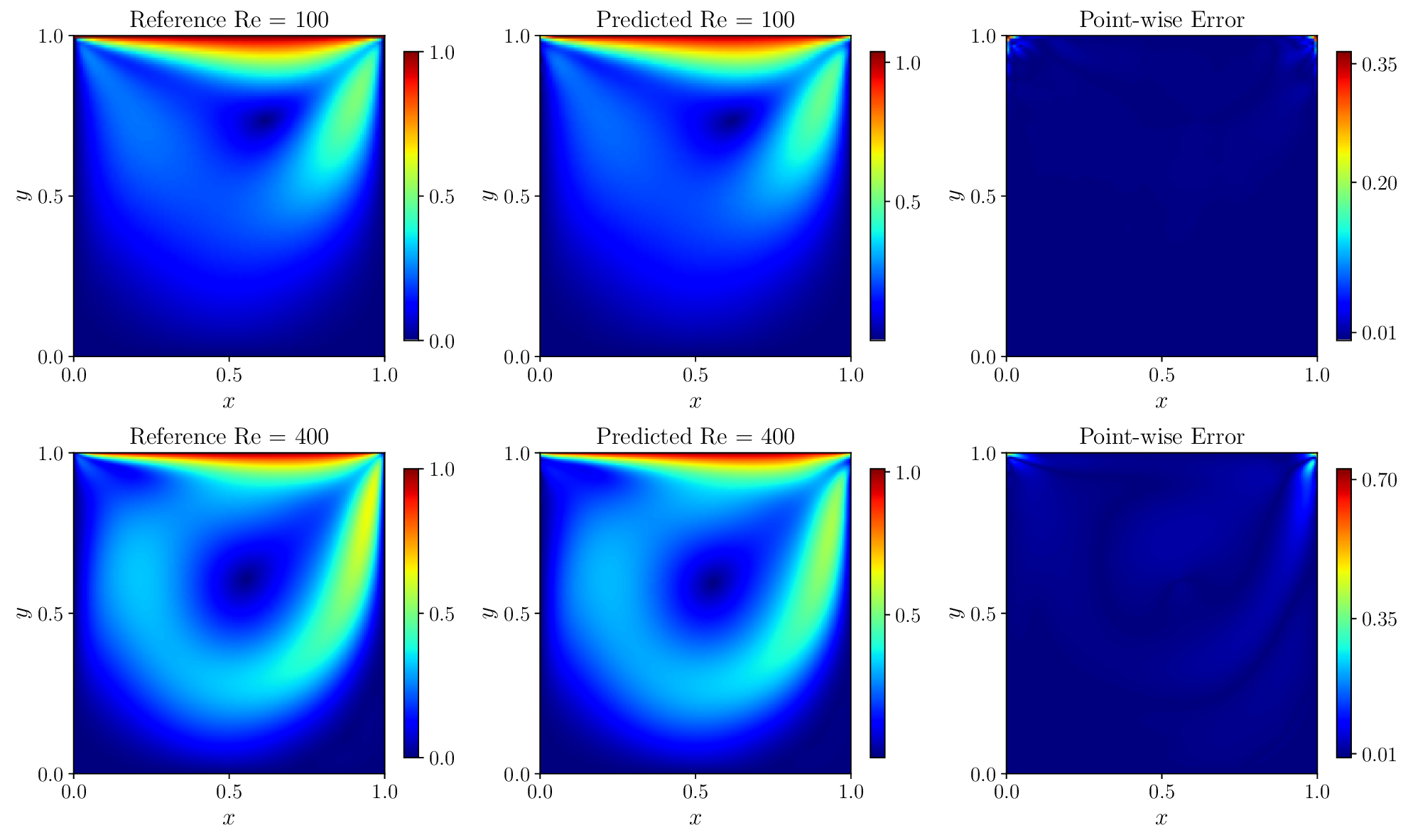}
\caption{{From left to right: The reference fluid flow, the W-PINN predicted fluid speed, and the point-wise absolute error for the steady-state lid-driven cavity flow \ref{eg:_new9} for Re=100 (top) and Re=400 (bottom).}}\label{fig:18_new}
\end{figure}

\begin{figure}[t!]
    \centering
    \includegraphics[width=0.8\linewidth]{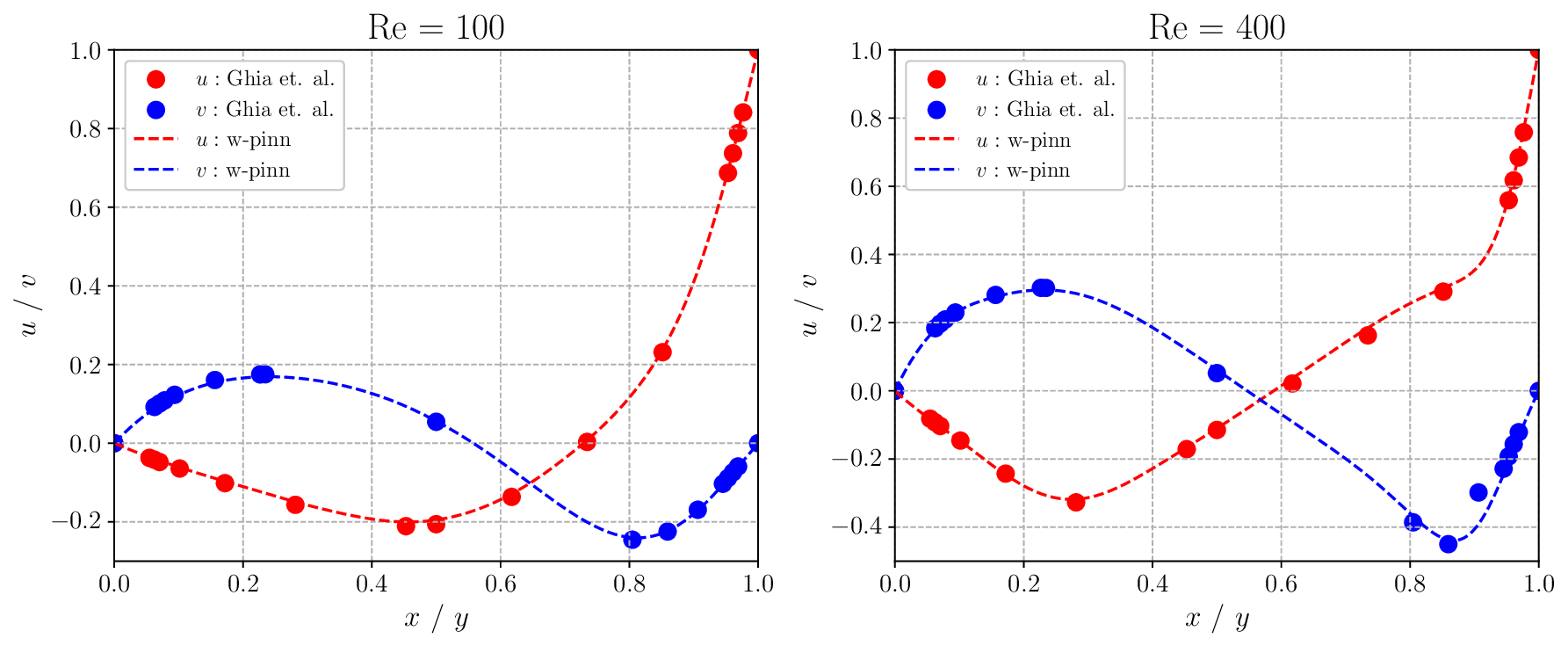}
\caption{{Comparison of W-PINN solution (dashed line) with Ghia's benchmark (scattered circles) for the steady-state lid-driven cavity flow \ref{eg:_new9} for Re=100 (left) and Re=400 (right). These data points lie on the horizontal and vertical lines through the geometric center of the cavity, for red color: $x=0.5$ and blue color: $y=0.5$}}\label{fig:19_new}
\end{figure}

{For this example, the network's parameters are tabulated in \ref{tab:17_new}. Figure \ref{fig:18_new} presents the predicted fluid flow speed distribution ($|V| = \sqrt{u^2+v^2}$), along with a comparison with the COMSOL reference solution, demonstrating successful capture of the characteristic vortex structure. The relative $L_2$-error for Re = 100 and Re = 400 are $2.75\pm1.02 \times 10^{-2}$ and $6.17\pm1.05 \times 10^{-2}$ respectively. W-PINN took about 20 minutes of training time for $10^4$ iterations on Nvidia A6000. At the higher Reynolds number of 400, where flow complexity increases, we uniformly sampled an additional $10^4$ collocation points from the upper left and upper right corners of the cavity (crossed region in Figure \ref{fig:cavity}). In addition, only four strategically chosen reference points (marked by red crosses in Figure~\ref{fig:cavity}) were added. Interestingly, these minimal additions reduced the relative $L_2$-error roughly by half. A validation against Ghia's benchmark data is illustrated in Figure \ref{fig:19_new}. These results demonstrate W-PINN's robust capability to handle complex fluid dynamics problems to a good extent.}
}

\section{Conclusion}\label{sec_concl}
We presented a wavelet basis function representation of PINN (W-PINN), and successfully demonstrated the advantages of W-PINNs over other PINN-based methods in addressing multiscale problems characterized by steep gradients, rapid oscillations, or singular behavior. The wavelet theory served as the foundation for the development of W-PINN, which combined the learning efficiency of PINN with the localization properties of wavelets (both in scale and space) to capture non-linear information within localized regions. A key advantage of the proposed framework is its elimination of the need for automatic differentiation (autograd) in computing residual derivatives, which substantially accelerates training compared to methods that rely on autograd. To support faster convergence behavior, we provide a comparative analysis of W-PINN using NTK theory. 

{Beyond the problems considered in this work, several promising directions emerge for future research. W-PINN can be naturally extended to parametric PDEs, as well as to inverse problems involving the identification of unknown source terms or governing parameters. The wavelet representation also enables the development of adaptive refinement strategies, in which the resolution level is dynamically increased in regions exhibiting sharp gradients or localized features. A limitation of the proposed method is its dependence on a careful selection of the wavelet basis for optimal performance. Another limitation is the exponential growth in the number of wavelet basis functions with increasing dimensionality, which leads to significant memory overhead and practical challenges in handling large wavelet matrices. Addressing these limitations through sparse representations, dimensionality reduction, or adaptive basis selection constitutes an important direction for future work. Nevertheless, the challenges targeted in this manuscript, namely, multiscale behavior, stiffness, and loss imbalance, are already sufficiently severe that conventional and many advanced PINN variants struggle to solve even in low-dimensional settings.} Furthermore, W-PINNs show strong potential for application to other classes of problems that exhibit inherent singular behavior, such as fractional differential equations. These models often present non-locality and memory effects, where the localized resolution capabilities of wavelets can offer a significant advantage.

Overall, the proposed W-PINN framework provides a robust and efficient alternative to existing physics-informed learning approaches for handling problems exhibiting multiscale behavior, and opens new possibilities for multi-scale representation in scientific machine learning. \\

\noindent \textbf{\large{CRediT authorship contribution statement}}\\

\noindent  \textbf{Himanshu Pandey:} Conceptualization, Methodology, Programming, Formal analysis, Writing - original draft.  \textbf{Anshima Singh:} Conceptualization, Investigation, Methodology, Formal analysis, Writing - original draft. \textbf{Ratikanta Behera:} Funding acquisition, Supervision, Writing - review \& editing.\\

\noindent \textbf{\large{Declaration of competing interest}}\\

\noindent The authors declare that there are no conflicts of interest. \\

\noindent \textbf{\large{Data Availability Statement}}\\

\noindent The data and source codes of problems addressed in this study are available at {\url{https://github.com/himanshup21/W-PINN.git}}\\

\noindent \textbf{\large{Acknowledgments}}\\

\noindent Ratikanta Behera is supported by the 
Anusandhan National Research Foundation (ANRF), Government of India, under Grant No. EEQ/2022/001065. We want to express our gratitude to the editor for taking the time to handle the manuscript and to the anonymous referees for their constructive comments.\\

% \noindent \textbf{\large{Preprint}}\\

% \noindent A preprint of this paper is available on arxiv with Ref. No. arXiv:2409.11847

\bibliography{References}	

\newpage
\appendix
\renewcommand{\thetable}{A.\arabic{table}} 
\setcounter{table}{0}   
\section{Hyperparameters}\label{A1}

\begin{table}[htbp]
\caption{Neural network and optimization hyperparameters for W-PINN}
\label{tab:hyperparams_net}
\small
\hspace{-1.2cm}
\begin{tabular}{p{7cm}|cccccc}
\hline
&&&&&\\
Problems & Depth & Width & Activation & Optimizer & Max iters \\
& {\footnotesize shallow+deep} & {\footnotesize neurons} &&&\\
& & {\footnotesize per layers} &&&\\
&&&&&\\
\hline
&&&&&\\
Example \ref{eg:1}~: 1D advection-diffusion  & 2~+~4 & 50 & $\tanh$ & Adam (lr: $10^{-5}$) & $5\times10^4$ \\

Example \ref{eg:2}~: Singularly perturbed nonlinear IVP & 2~+~4 & 50 & $\tanh$ & Adam (lr: $10^{-5}$) & $10^5$ \\

Example \ref{eg:3}~: Singularly perturbed nonlinear BVP & 2~+~6 & 50 & $\tanh$ & Adam (lr: $10^{-4}$) & $10^4$ \\

Example \ref{eg:4}~: FHN model & 3~+~7 & 100 & $\tanh$ & Adam (lr: $10^{-5}$) & $2\times10^4$ \\

Example \ref{eg:5_new}~: Heat conduction with large gradients  & 2~+~7 & 50 & $\tanh$ & Adam (lr: $10^{-4}$) & $2\times10^4$ \\

Example \ref{eg:6_new}~: 2D Helmholtz with high-frequency & 2~+~4 & 50 & $\tanh$ & Adam (lr: $10^{-5}$) & $5\times10^4$ \\

Example \ref{eg:7_new}~: Allen--Cahn with periodic BCs  & 2~+~4 & 100 & $\tanh$ & Adam (lr: $10^{-5}$) & $10^5$ \\

Example \ref{eg:8_new}~: Maxwell's Equation & 2~+~6 & 50 & $\tanh$ & Adam (lr: $10^{-4}$) & $3\times10^4$ \\

Example \ref{eg:9_new}~: Lid-driven cavity & 3~+~9 & 100 & $\tanh$ & Adam (lr: $10^{-5}$) & $10^4$ \\
&&&&&\\
\hline
\end{tabular}
\end{table}

\begin{table}[htbp]
\caption{Resolution and sampling parameters for W-PINN}
\label{tab:hyperparams_sampling}
\small
\hspace{-1.2cm}
\begin{tabular}{l|cccccc}
\hline
&&&&&\\
Problems & Resolution~($J$) & $\delta$ & $N_r$ & $N_{bc}$ & $N_{ic}$ \\
&&&&&\\
\hline
&&&&&\\
Example \ref{eg:1}~: 1D advection-diffusion  
& $J_M,J_G,J_{\mathcal{M}}= [0,8],[0,10],[0,9]$ & 1.0 & $10^3$ & -- & -- \\

Example \ref{eg:2}~: Singularly perturbed nonlinear IVP
& $J_M,J_G,J_{\mathcal{M}} = [0,9],[0,10],[0,10]$ & 1.0 & $10^4$ & -- & -- \\

Example \ref{eg:3}~: Singularly perturbed nonlinear BVP
& $J_M,J_G = [0,8],[0,9]$ & 1.0 & $10^4$ & -- & -- \\

Example \ref{eg:4}~: FHN model
& $J_M,J_G = [0,10],[0,11]$ & 1.0 & $10^4$ & -- & -- \\

Example \ref{eg:5_new}~: Heat conduction with large gradients
& $J_x=J_t=[-6,5]$ & 0.2 & $10^4$ & $10^3$ & $5\times10^2$ \\

Example \ref{eg:6_new}~: 2D Helmholtz with high-frequency 
& $J_x=J_y=[-4,5]$ & 0.2 & $10^4$ & $10^3$ & -- \\

Example \ref{eg:7_new}~: Allen--Cahn with periodic BCs
& $J_x=[-5,6],\ J_t=[-5,5]$ & 0.4 & $2\times10^4$ & $2\times10^3$ & $10^3$ \\

Example \ref{eg:8_new}~: Maxwell's Equation 
& $J_x=J_t=[-5,5]$ & 0.5 & $10^4$ & $10^3$ & $5\times10^2$ \\

Example \ref{eg:9_new}~: Lid-driven cavity
& $J_x=J_y=[-15,5]$ & 0.5 & $4\times10^4$ & $2\times10^3$ & -- \\

&&&&&\\
\hline
\end{tabular}
\end{table}

\end{document}